\documentclass{ifacconf}

\usepackage{graphicx}      
\usepackage{natbib}        
\usepackage{xcolor}

\usepackage{amsmath}
\usepackage{amssymb}
\usepackage{subcaption}
\usepackage{url}

\usepackage{microtype}

\usepackage{titlesec}

\DeclareMathOperator*{\argmax}{\arg\max}   

\begin{document}
\begin{frontmatter}

\title{Varying Bundle Size Reactive Multi-Task Assignment using Selective Cost Estimation for Multi-Agent Systems}

\thanks{This work was supported by the European Union’s Horizon Europe programme under the SUNRISE-6G project (Grant Agreement No. 101139257).}


\author[First]{Niklas Dahlquist}
\author[First]{Shridhar Velhal} 
\author[First]{George Nikolakopoulos}

\address[First]{Luleå University of Technology, Luleå, SE-971 87 Sweden (e-mail: niklas.dahlquist@ltu.se).}

\begin{abstract}                
This paper presents a scalable framework for multi-robot task allocation in complex environments where estimating task execution costs is computationally expensive. While combinatorial auction-based approaches offer reliable solutions, the exponential complexity of bundle generation typically renders them intractable for real-time reactive applications, particularly when accurate path planning is required for cost validation. We address this through a distributed, two-stage multi-fidelity bundle generation approach. Agents utilize a local search tree guided by a low-fidelity heuristic (such as euclidean distance) to rapidly explore the bundle space, applying high-fidelity path planning only to the most promising candidates in a best-first manner. These refined bids are then submitted to a central coordinator that solves a set packing problem to ensure global feasibility and maximize the overall utility. Simulation results in multiple environments demonstrate that the framework is able to improve the performance of reactive auction-based task allocation. Overall, the presented framework is shown to enable reactive task allocation with dynamic bundle sizes in multiple settings without exposing the agents' state and internal cost estimation models.
\end{abstract}

\begin{keyword}
Reactive task allocation, multi-agent coordination, combinatorial optimization
\end{keyword}

\end{frontmatter}

\section{Introduction}
Modern multi-agent systems face an increasingly dynamic and complex landscape. Tasks can arrive without a priori information, in batches or streams (Dahlquist et al., 2023), rewards can be time-sensitive, and the cost of task execution depends on geometry, obstacles, and agent specifications. Classical task assignment methods \cite{CHAKRAA2023104492} provide an efficient way to enforce non-overlap between tasks and capacity constraints for individual agents, but they implicitly assume that bundle (ordered set of tasks) values (i.e., the total utility of assigning a set of tasks to an agent) are known or cheap to evaluate. In multiple emerging applications such as: last-mile delivery~\cite{OSTERMEIER2023680}, routine inspection~\cite{LEE2023100018}, emergency response \cite{GHASSEMI2022103905}, or collaborative exploration \cite{AZPURUA2023104304} task utility estimation can be computationally expensive. Estimating the utility of executing any ordered sequence of tasks many times requires solving a nontrivial motion-planning problem. Consequently, searching the enormous space of potential bundles quickly becomes infeasible. 

While fully centralized assignment approaches can enforce global feasibility and optimality \cite{CHAKRAA2023104492}, such solutions often struggle to scale when every bundle evaluation requires computationally heavy planning and can ignore the fact that agents often possess private information (such as local maps and model specific information) needed to estimate accurate execution costs. At the same time, it is often desirable to allow agents to be as anonymous as possible so that internal cost estimation models are not required to be shared globally. Fully decentralized schemes (e.g., consensus-based bundle building \cite{cbba2009}), on the other hand, can achieve a high degree of scalability but can suffer from suboptimal solutions and limited global constraint handling. The separation between these concerns highlights the potential benefits of distributed task-allocation protocols where agents can leverage their local knowledge, while keeping a high degree of anonymity, for estimating the cost related to executing certain sets of tasks; while a minimal global coordinator distributes tasks and ensure fairness between all agents.


In this work, we propose a two-stage, multi-fidelity bundling approach to improve upon two important challenges related to combinatorial task allocation.
We address both the limited scalability in optimizing the assignments of available tasks and the sub-problem of estimating task execution costs for the available tasks; since both issues stems from the issue that the number of potential task bundles (ordered sets of tasks) scales factorially with the number of available tasks. The initial stage of the proposed bundling approach uses a computationally cheap heuristic (euclidean proximity) to rapidly explore a candidate task tree and generate many potential task bundles for each agent. The second stage then performs a high-fidelity computation selectively, based on a best-first refinement process to ensure that the computation is only performed for the most promising bundles in the task tree, to estimate realistic costs for a set of potential task bundles. This pipeline initially exposes a large and diverse candidate set at a low computational cost, and limits the expensive planning to a subset of task bundles that more meaningfully influences the final task allocation.

\subsection{Background}

Multiple surveys on multi-robot task allocation formalize task/agent taxonomies, coordination schemes, and optimality notions \citep{Gerkey2004}. Centralized methods, such as optimization based methods \cite{CHAKRAA2023104492} or classic auction based methods \cite{Dias-2004-8847}, enable strong global performance guarantees and can handle constraints well. But often rely on bundle/task execution values that are expensive to compute. Decentralized approaches, such as Consensus-Based Bundle Algorithms and their variants, scale well and provide conflict resolution via consensus \citep{cbba2009}. Such approaches work well when costs are cheap and fast to estimate and communication is readily available but the requirement of full connectivity/consensus could break the privacy/bandwidth constraints posed in some scenarios. When estimating realistic bundle costs requires extensive planning continuous re-estimation of the cost can limit the overall reactivity. 

Optimizing the allocation of task-bundles is in general a NP-hard problem \citep{SANDHOLM20021}. Multiple works focus on winner determination given fixed bids but fewer works focus on how to generate bundles efficiently when estimating their values are computationally expensive. This work focuses on the distributed part of the problem, where a two-stage, multi-fidelity bundle generation approach is presented to separate cheap breadth search from an expensive accurate cost estimation.

Relative to fully decentralized methods, we keep local high-fidelity cost estimations (agents are given the authority to compute their own bundle-specific costs) but perform a central assignment optimization via a set-packing integer program to handle global conflicts, preserving anonymity of the agents (no internal models are shared) and enables strong global consistency.

\subsection{Contributions} \label{sec:contributions}
The following are the main contributions of this article. i) A variable bundle-size reactive task-assignment approach is proposed for tasks without a priori information. By adjusting the bundle size and the considered bundles the proposed method aims to improve the performance of reactive task assignment by more efficiently assigning multiple tasks to every agent through a centralized set-packing optimization with diminishing bundle rewards. ii) A two-stage approach is proposed, in which a depth-limited, variable-width beam search using a euclidean distance as a heuristic is utilized to create a search tree, and a local high-fidelity cost is computed only for the k-best branches to reduce the computational burden of path computations. Also, in bidding, only the accumulated bundle cost is shared with the centralized system allowing individual task cost models and agent state to remain private. iii) We present a simulation-based scalability analysis of the approach in multiple different environments (inspired by forests, natural caves and tunnel systems) to approach real-world applicability. We evaluate all computational stages of the full task allocation architecture, from global optimization to local bundle-construction and cost estimation.

\section{Problem Formulation} \label{sec:formulation}
In this section, we present a concise description of the combinatorial auction-based multi-agent task allocation framework, the notations used, and ending with formulating the main problem formulation of this work.

In order to describe the combinatorial task allocation architecture, we begin by defining the necessary notations. In a multi-agent setting, the set of available tasks, at a specific time instant \(t\), is denoted by \(\mathfrak{T}_t = \{ T_1, T_2, \dots, T_{n_t} \} \), where \(n_t \in \mathbb{N}\) is the number of tasks available at that time. A team of agents \(\mathfrak{R} = \{ R_1, R_2, \dots, R_{n_a} \}\) are available for executing the tasks, where \(n_a \in \mathbb{N}\) denotes the number of agents. The bundles  \(\mathfrak{B}_{i, j}\) are defined as ordered sets of tasks that agent \(R_i\) can execute. We consider the scenario where the full set of tasks are not known a priori, meaning that the set \(\mathfrak{T}_t\) varies with time as new tasks are introduced or as tasks are executed by the agents.

We consider an auction based task allocation framework consisting of distributed computation of cost estimates and a light central coordinator for optimizing the current task allocation. The framework is made up of three discrete stages that are executed iteratively to reactively assign tasks during dynamic scenarios. The three stages are the following: 
i) Announcement stage: The central coordinator tracks the currently available tasks and periodically announces the set of available tasks \(\mathfrak{T}_t\) to all agents \(\mathfrak{R}\).
ii) Bidding stage: Every agent computes the costs (\(c_{i,j}\)), for agent \(R_i\) to execute bundle \(\mathfrak{B}_{i, j}\). The costs are submitted back to the coordinator as bids. This approach gives the agents the local autonomy to compute the costs through path-planning over a known map of the environment without sharing any state-information with the centralized coordinator. 
iii) Task allocation stage: To decide the optimal task allocation, the centralized coordinator announces the assigned bundles of tasks selected by solving the following set-packing problem: 
\begin{equation}
    \begin{aligned}
         x^* = & \argmax_{x_{i, j}} \sum_{(i,j) | b_{(i, j)} \in B} ( R_{i,j} - c_{i,j}) \cdot x_{i,j} \\
        \textrm{s.t.} \quad & \sum_{(i,j) | b_{(i, j) \in B}} b_{i, j, k} \cdot x_{i, j} \leq 1, \text{ }  \forall k \in \{1, \ldots, n_t\} \\
        & \sum_{(i,j) | b_{(i, j)} \in B} x_{i, j} \leq 1, \text{ }  \forall i \in \{1, \dots, n_a\} \\
        & x_{i, j} \in \{0, 1\} \textrm{ } \forall (i, j) \\
        & b_{i, j, k} \in \{0, 1\} \textrm{ } \forall (i, j, k)
    \end{aligned}
    \label{eq:max_optimization_constrained}
\end{equation}

where \(x_{i, j}\in\{0,1\}\) is an assignment variable indicating if agent \(R_i\) is assigned to bundle \(B_{i, j}\) and the assignment vector \(x^\star\in\{0,1\}^{n_a \times n_b}\), where \(n_b\) denotes the number of available bundles, defines the optimal distribution of tasks \(\mathfrak{T}_t\) among the agents \(\mathfrak{R}\). \(R_{i, j}\in\mathbb{R}\) represents the `reward' for agent \(R_i\) to complete bundle \(B_{i, j}\). \(B\) is the set of two-tuples representing the edges of a bipartite graph between the set of agents and the set of tasks-bundles. A key challenge of this task allocation framework is that the number of bundles \(B_{i, j}\), that are required for every \(R_i\), grows factorially with the number of available tasks \(n_a\).

\emph{Problem definition:} As the number of tasks, agents and the scale of the environment grows large, computing the execution costs for the full set of potential task bundles quickly becomes infeasible. In this article, we will explore a scalable, two-stage bundle construction and task-cost estimation procedure that explores a large and diverse subset of the bundle space using cheap, low-fidelity heuristic cost estimates and then selectively, using a best-first search, computes the high-fidelity costs \(c_{i, j}\) for a subset of the candidate bundles. This is to be done while preserving the agent anonymity and keeping individual task costs and local estimation models private while maintaining computational tractability and reactivity as \(n_t\) and \(n_a\) grows large.

\section{Methodology} \label{sec:methodology}

This section details how the candidate bundles are selected based on a cheap euclidean heuristic to generate a tree of potential task orders and how a subset of ordered task-sets, bundles, are selected and refined with a high-fidelity cost estimation. The refined bundles are then utilized to select the set of tasks to assign to the individual agents.

\subsection{Centralized Optimization}

During the task allocation stage, as detailed in Sec. \ref{sec:formulation}, the centralized optimization problem \eqref{eq:max_optimization_constrained} has to be solved to provide the optimal task allocation \(x^*\). To preserve the anonymity of the internal cost estimation models of the agents, account for uncertainty of future task assignments, and also balance the assigned bundle sizes to encourages fairness between agents a diminishing reward is introduced based on the bundle size. Every task is considered to have a reward \(R_{t}\) and the total reward for bundle \(\mathfrak{B}_{i, j}\) can then be defined as
\(
    R^{\mathrm{tot}}_{i, j} = \sum_{l \in \mathfrak{T}_l | T_l \in \mathfrak{B}_{i, j}} \alpha^l R_t
\)
where \(\alpha < 1\) is a selectable parameter that defines how much the reward of additional tasks contributes to the total bundle reward. This can be also expressed as a geometric sum
\(
     R^{\mathrm{tot}}_{i, j} = R_t \cdot \frac{1 - \alpha^{|\mathfrak{B}_{i, j}|}}{1 - \alpha}
     \label{eq:geometric_sum_objective_diminishing_reward}
\)
where \(|\mathfrak{B}_{i, j}|\) defines the total amount of tasks in bundle \(\mathfrak{B}_{i, j}\). This ensures that the objective function is affected by the bundle reward, since only the total bundle cost estimation \(c_{i, j}\) is known. The objective function in \eqref{eq:max_optimization_constrained} can thus be augmented as:
\begin{equation}
     x^* = \argmax_{x_{i, j}} \sum_{(i,j) | b_{(i, j)} \in B} ( R^{\mathrm{tot}}_{i, j} - c_{i,j}) \cdot x_{i,j}.
\end{equation}

\subsection{Bid Generation}

To manage the factorial complexity of bundle generation and cost estimation with regard to the amount of available tasks, we decompose the bidding process into two stages: a fast candidate tree generation via a heuristic depth-limited, variable-width beam search, and a high-fidelity cost estimation via path planning using a best-first search based on the candidate tree.

\subsubsection{Candidate Tree Generation:}

Each agent \(R_i\) builds a rooted tree \(\mathcal{T_\textrm{}} = (\mathcal{N}, \mathcal{V})\), where \(\mathcal{N}\) denotes the candidate tasks and \(\mathcal{V}\) defines the connections, to explore the space of feasible bundles (task sequences). Every node \(n_i \in \mathcal{N}\) in the search tree represents a partial candidate bundle \(b_i\) containing all the ancestor nodes of \(n_i\). A beam search strategy is applied to generate the candidate tree \(\mathcal{T}\). At each depth \(D\), only the top \(K_D\) (a variable search width) most promising next tasks are for added for expansion using an euclidean distance heuristic, chosen for computational speed, meaning that the children \(C_i\) are selected as
\begin{equation}
    C_i = \arg\min_{T^p \in \mathfrak{T}_t^p} d_T(n_i,K_D)
\end{equation}
where \(d_T(n_i, K_D)\) is the sum of distances from the location of \(n_i\) to the locations of the \(K_D\) chosen candidate children nodes \(C_i\) and \(\mathfrak{T}_t^p = \{\mathfrak{T}_t \setminus b_i \}\) denotes the set of currently potential next tasks for \(n_i\). The candidate tree is then iteratively expanded until the target depth \(D_T\) is reached.

\subsubsection{Bundle Cost Estimation:}

To bridge the gap between the computational efficiency of the heuristic tree search and the accuracy required for valid task assignment, a best-first selection strategy coupled with a high-fidelity cost refinement is introduced. The search tree \(\mathcal{T}\) contains numerous potential task bundles, and evaluating the true execution cost \(C_{true}(b)\) for every node in \(\mathcal{T}\) is computationally prohibitive since it requires solving the motion planning problem for each bundle. Instead, the total euclidean distance, calculated during the tree generation, is utilized as an initial cost estimate to rank potential bundles in a priority queue. This allows the agent to iteratively extracts the top-promising bundle from the queue for calculating the true cost \(C_{true}(b)\). Every time a bundle cost is estimated the heuristic euclidean cost can be updated with \(C_{true}(b)\) to make sure that the priority queue stays relevant. This makes the agent focus its limited computational budget on the bundles that are most likely to be competitive. In this work the \(K_b\) number of top-promising bundles are calculated.

To further reduce computational load, since the cost of traversing between two task locations is invariant across different bundles, calculated path segments are cached in a local lookup table. If a subsequent candidate bundle requires a transition from \(T_i\) to \(T_j\) that has already been computed, the planner retrieves the stored cost, bypassing the expensive motion planning step.

\begin{figure*}[h!]
    \centering
    \begin{subcaptionblock}[c]{0.3\textwidth}
        \centering
        \includegraphics[width = \textwidth]{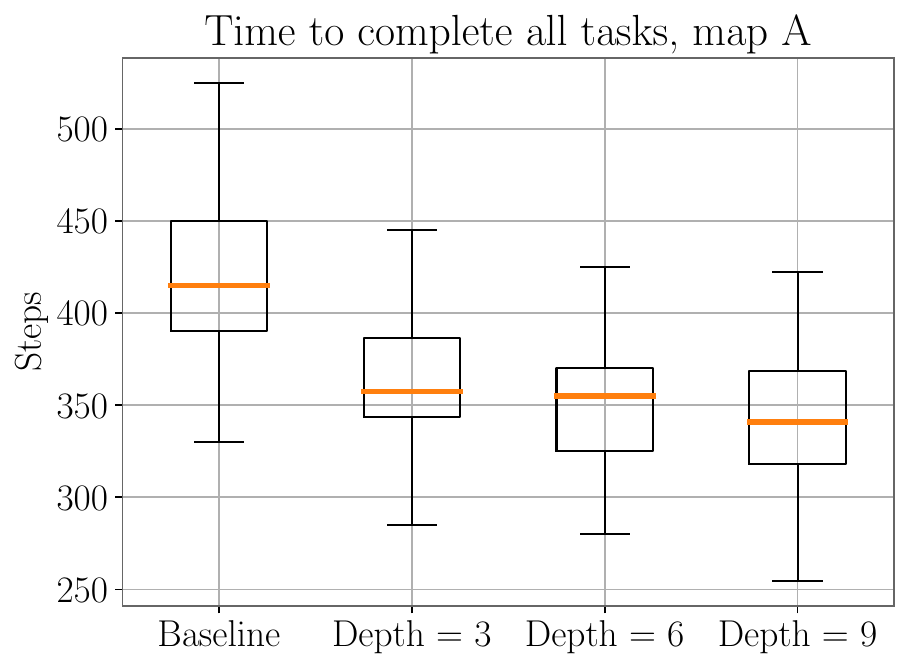}
        \caption{}
    \end{subcaptionblock}%
    \hfill
    \begin{subcaptionblock}[c]{0.3\textwidth}
        \centering
        \includegraphics[width = \textwidth]{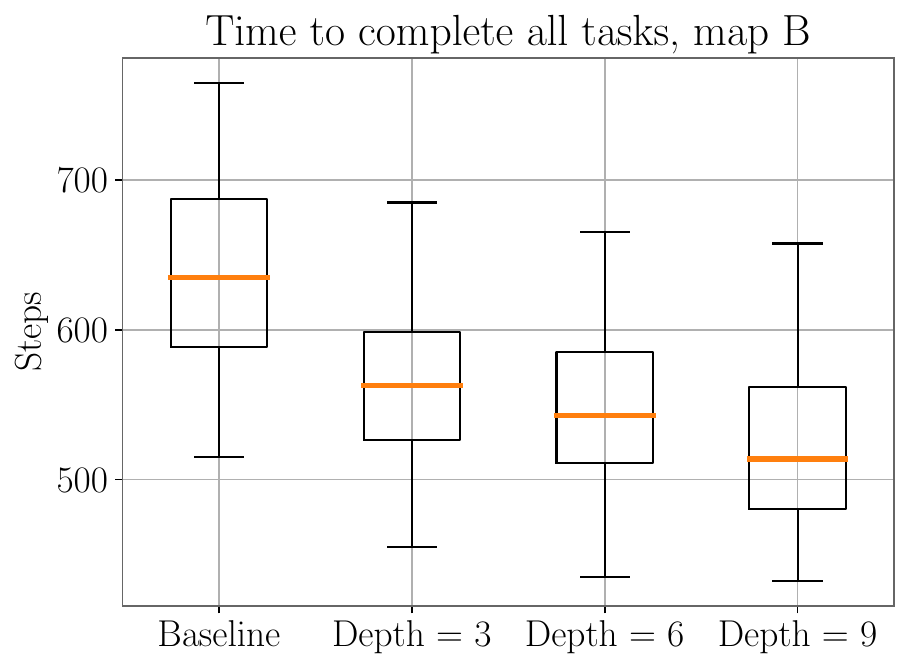}
        \caption{}
    \end{subcaptionblock}%
    \hfill
    \begin{subcaptionblock}[c]{0.3\textwidth}
        \centering
        \includegraphics[width = \textwidth]{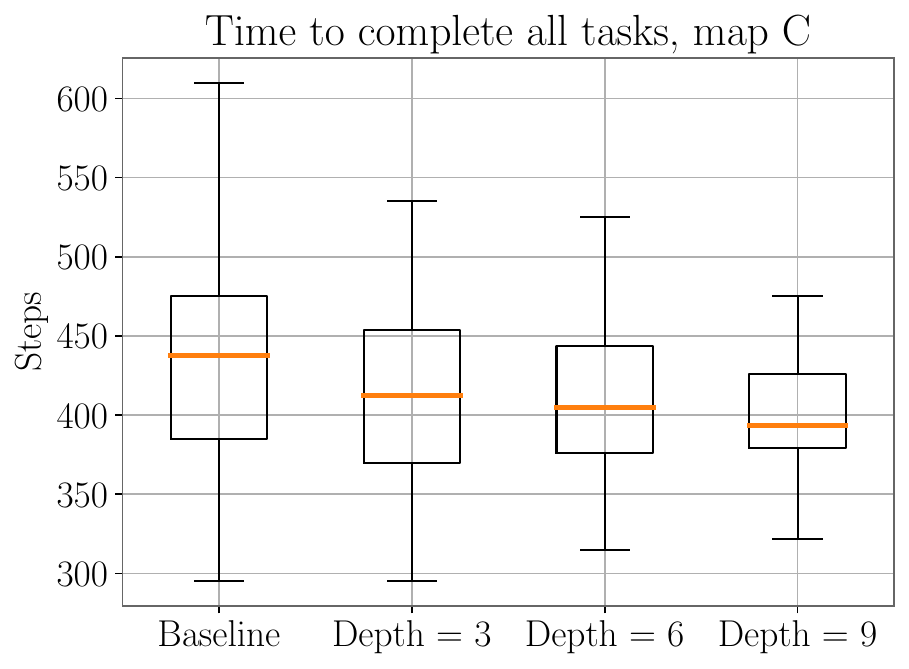}
        \caption{}
    \end{subcaptionblock}%

    \caption{Results for scenario 1. Total time required to execute all tasks for the three evaluation maps.}
    \label{fig:total_steps_static}
\end{figure*}

\begin{figure}[t]
    \centering
    \begin{subcaptionblock}[c]{0.31\columnwidth}
        \centering
        \includegraphics[width = \textwidth]{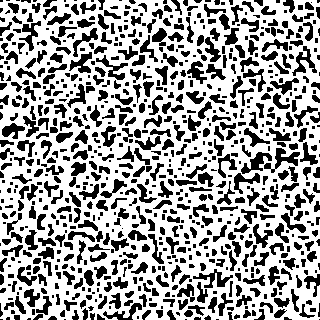}
        \vspace{0.5pt}
        \caption{Map A.}
        \label{fig:three_maps:a}
    \end{subcaptionblock}%
    \hfill
    \begin{subcaptionblock}[c]{0.31\columnwidth}
        \centering
        \includegraphics[width = \textwidth]{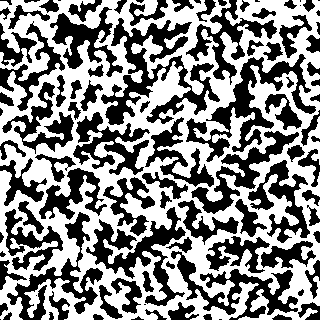}
        \vspace{0.5pt}
        \caption{Map B.}
        \label{fig:three_maps:b}
    \end{subcaptionblock}%
    \hfill
    \begin{subcaptionblock}[c]{0.31\columnwidth}
        \centering
        \includegraphics[width = \textwidth]{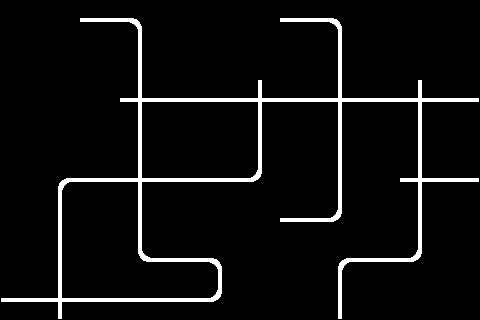}
        \vspace{0.5pt}
        \caption{Map C.}
        \label{fig:three_maps:c}
    \end{subcaptionblock}%
    
    \caption{The maps used for evaluation. (a) a map with smaller, scattered obstacles, (b) a map with large complex obstacles, and (c) a tunnel-like map.}
    \label{fig:three_maps}
\end{figure}

\begin{figure}[tb]
    \centering

    \begin{subcaptionblock}[c]{0.33\textwidth}
        \centering
        \includegraphics[width = \textwidth]{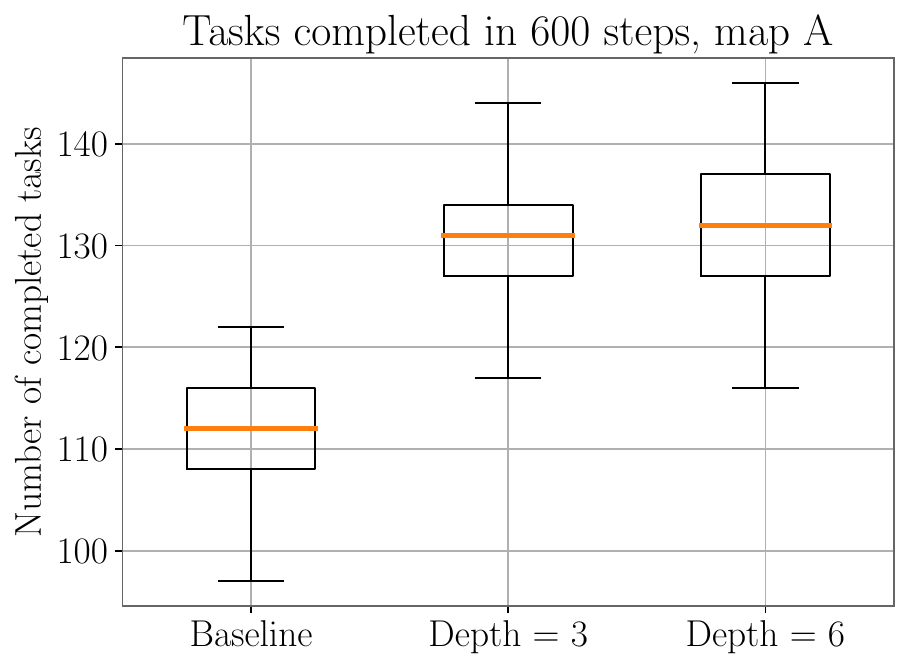}
        \caption{}
    \end{subcaptionblock}%
    \\
    \begin{subcaptionblock}[c]{0.33\textwidth}
        \centering
        \includegraphics[width = \textwidth]{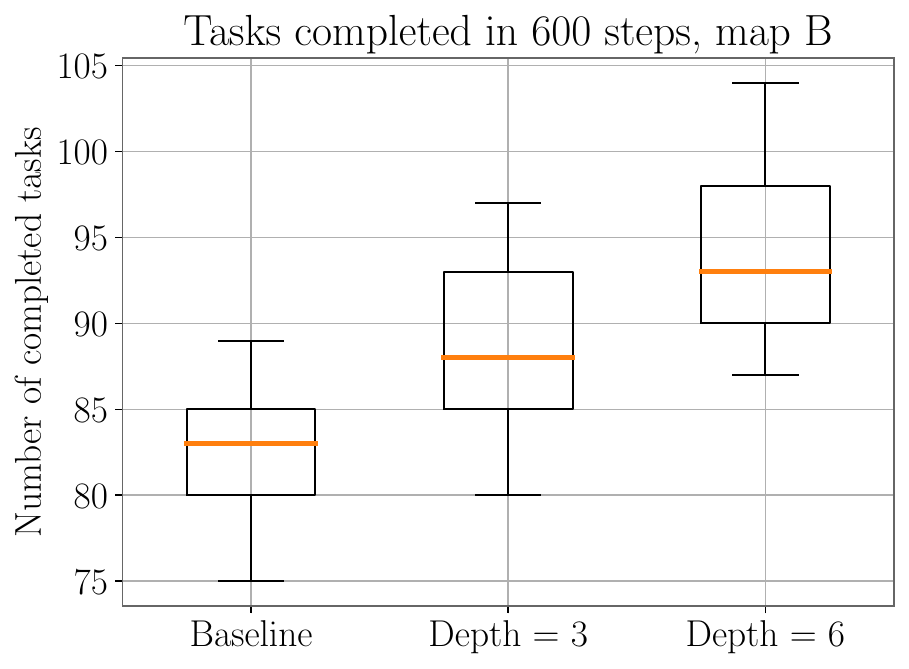}
        \caption{}
    \end{subcaptionblock}%

    \caption{Results for scenario 2.}
    \label{fig:total_steps_dynamic}
\end{figure}

\section{Simulation Results} \label{sec:results}

To evaluate the performance of the proposed framework, three different environments are considered. The maps of the environments can be seen in Fig. \ref{fig:three_maps} and have a moderately large size of 320x320 for map A and map B, while map C has a size of 480x320. The first map, shown in Fig. \ref{fig:three_maps:a}, illustrates a relatively open environment with smaller obstacles inspired by outdoor environments (such as forests). The second map, shown in Fig. \ref{fig:three_maps:b}, consists of more complex and larger non-convex obstacles and is inspired by typical cave environments. The third map, shown in Fig. \ref{fig:three_maps:c}, is based on a tunnel layout. The optimization function (and bid generation) is executed every \(5\) time steps and the agents are modeled as point-masses that can move with a \(1\) pixels per time step, every scenario is computed \(100\) times to generate the statistics and plots presented, and all the specific parameters used in the evaluation scenarios are summarized in Table \ref{tab:scenario_parameters}. 

\begin{table}[t]
    \centering
    \caption{The parameters used to throughout the evaluation of the proposed framework.}
        \begin{tabular}{|c|c|}
            \hline
            Description & Value \\
            \hline
            Reward per task & \(R_t = 300\) \\
            Diminishing reward factor & \(\alpha = 0.7\) \\
            Variable tree width & \(K_D = [6, 5, 4, 3, 3, 3, 2, 1, 1]\) \\
            Candidate tree depth & \(D_T = [3, 6, 9]\) \\
            Number of bundles & \(K_b = 1500\) \\
            Penalty factor & \(\beta = 0.5\) \\
            Penalty scaling & \(\gamma = 0.005\) \\
            \hline
        \end{tabular}
    \label{tab:scenario_parameters}
\end{table}

The evaluation is performed on a computer with an AMD Ryzen 7 PRO 5850U, 32 GB RAM and running Ubuntu 24.04. The path planning is based on a \(\text{D}^*\) algorithm and implemented using C++ and the optimization problem \eqref{eq:max_optimization_constrained} is defined using \cite{ortools} and solved using SCIP \cite{GamrathEtal2020OO}.

\subsection{Evaluation Scenarios}
The proposed framework, detailed in section \ref{sec:methodology}, is evaluated in three different scenarios. 1) Static performance for completing all available tasks. 2) Dynamic performance in a scenario where tasks are introduced continuously during execution and the mission is abruptly aborted. 3) A use-case where a team of agents has to complete an `inspection-style' mission. 

\begin{figure*}[tb]
    \centering
    \begin{subcaptionblock}[c]{0.25\textwidth}
        \centering
        \includegraphics[width = \textwidth]{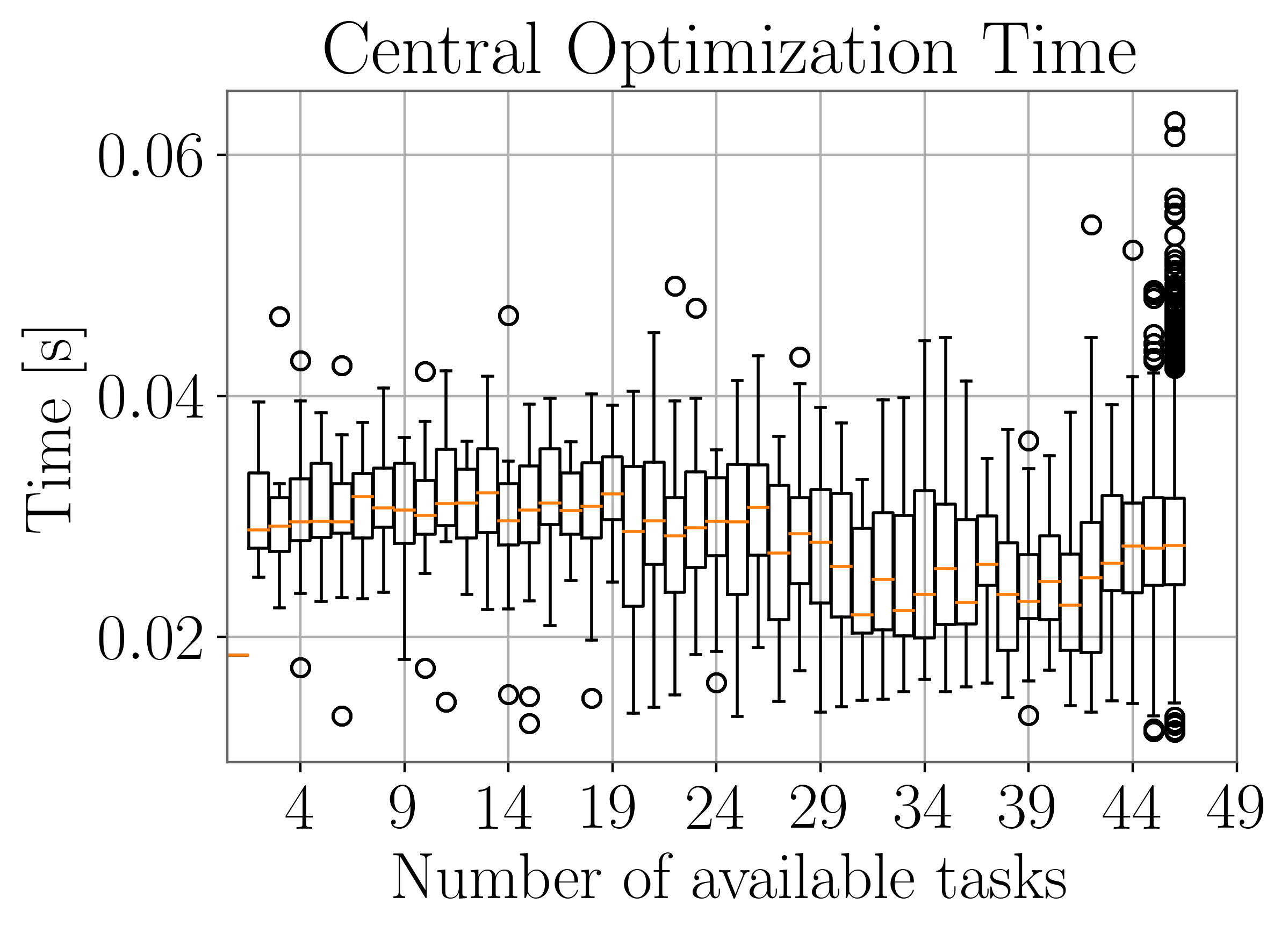}
        \caption{}
    \end{subcaptionblock}%
    \hfill
    \begin{subcaptionblock}[c]{0.25\textwidth}
        \centering
        \includegraphics[width = \textwidth]{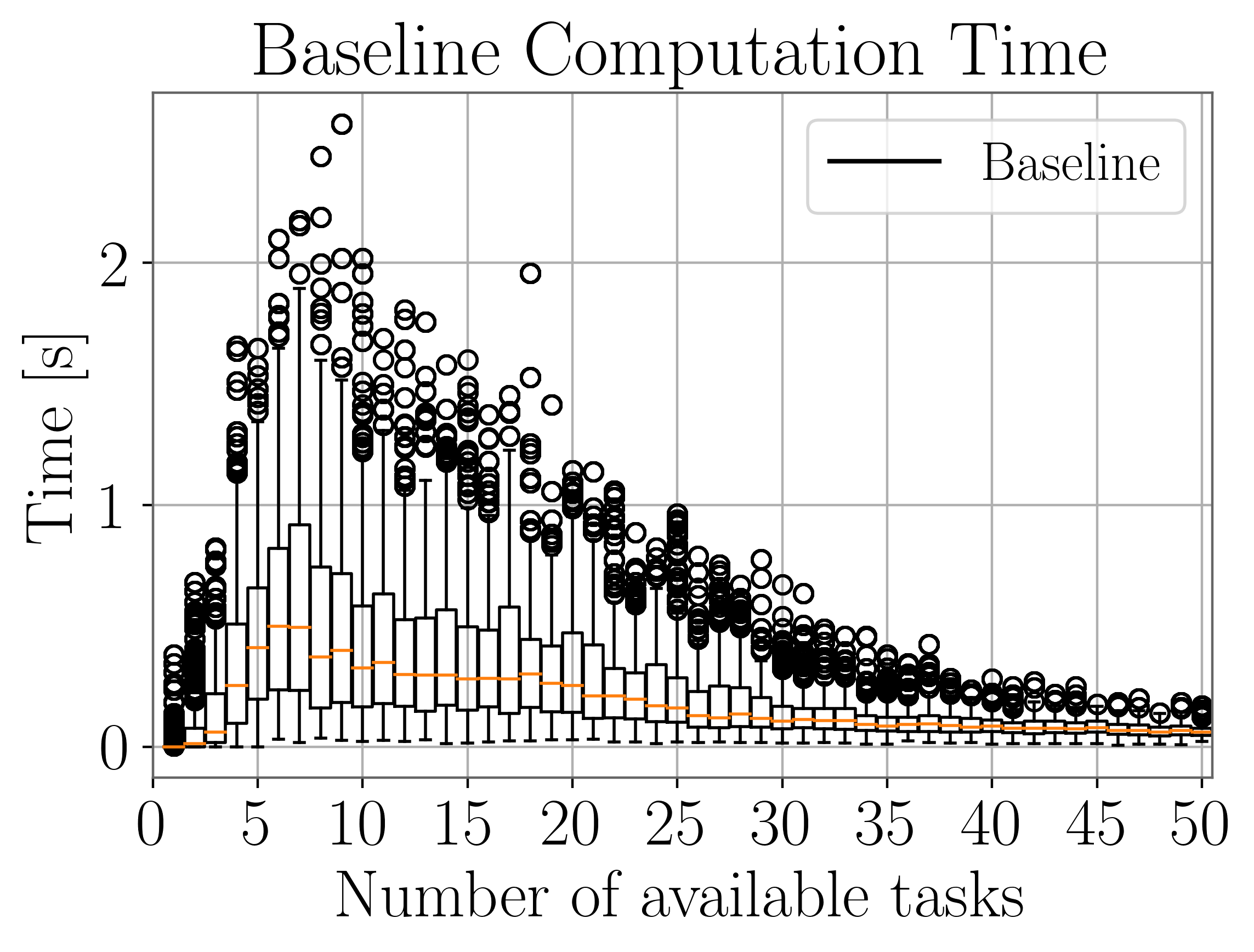}
        \caption{}
    \end{subcaptionblock}%
    \hfill
    \begin{subcaptionblock}[c]{0.25\textwidth}
        \centering
        \includegraphics[width = \textwidth]{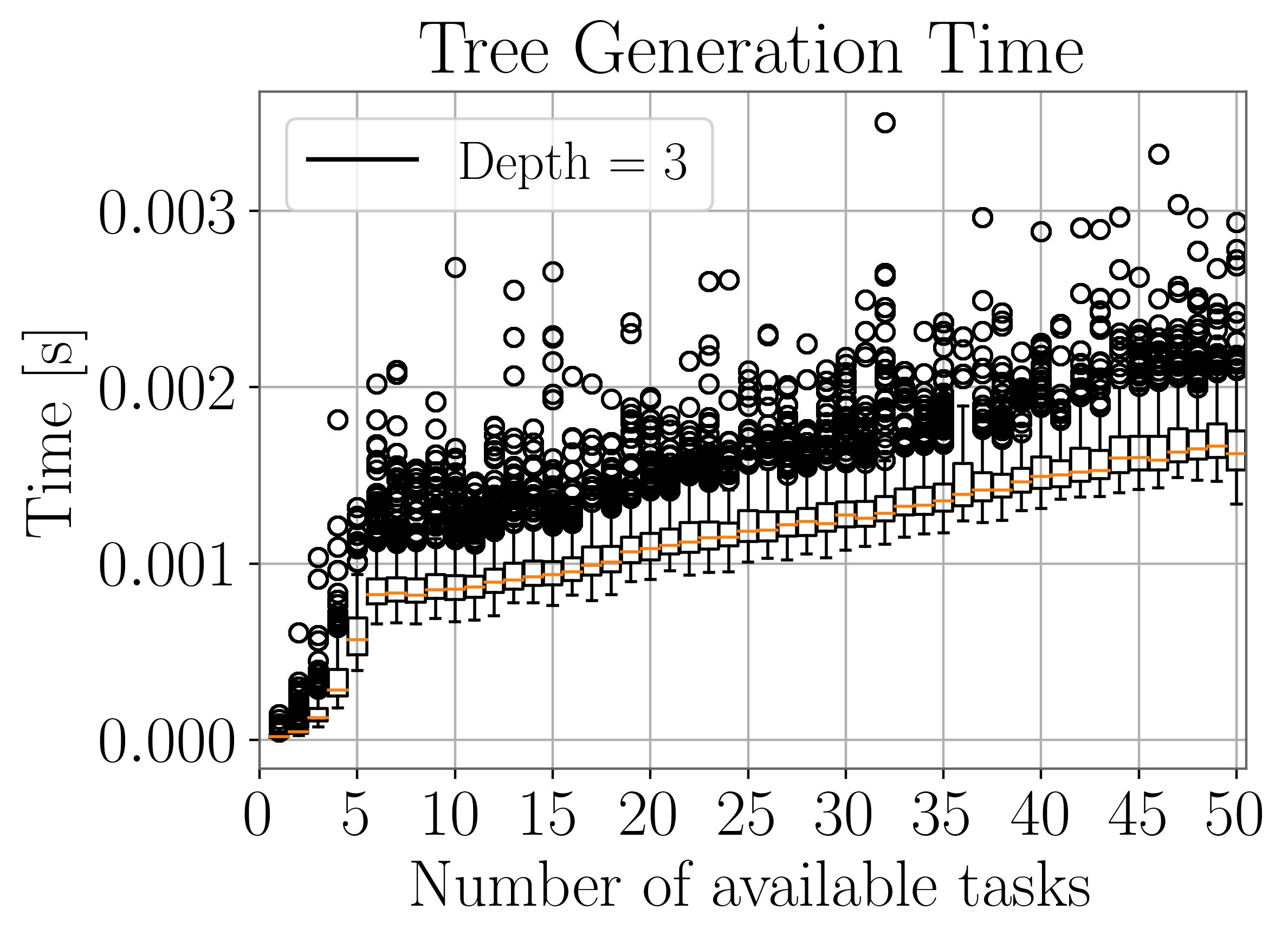}
        \caption{}
    \end{subcaptionblock}%
    \hfill
    \begin{subcaptionblock}[c]{0.25\textwidth}
        \centering
        \includegraphics[width = \textwidth]{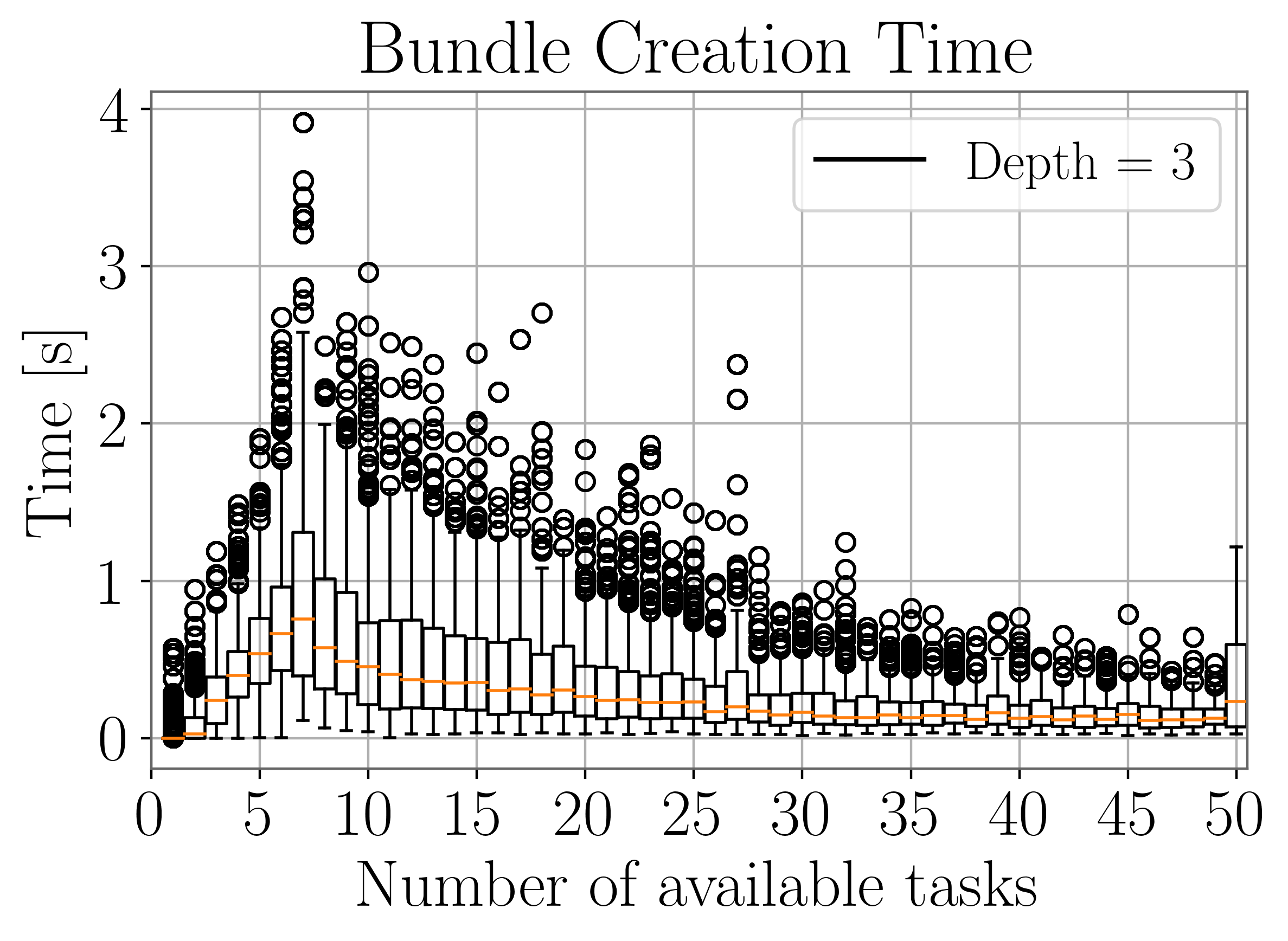}
        \caption{}
    \end{subcaptionblock}%
    \\
    \begin{subcaptionblock}[c]{0.25\textwidth}
        \centering
        \includegraphics[width = \textwidth]{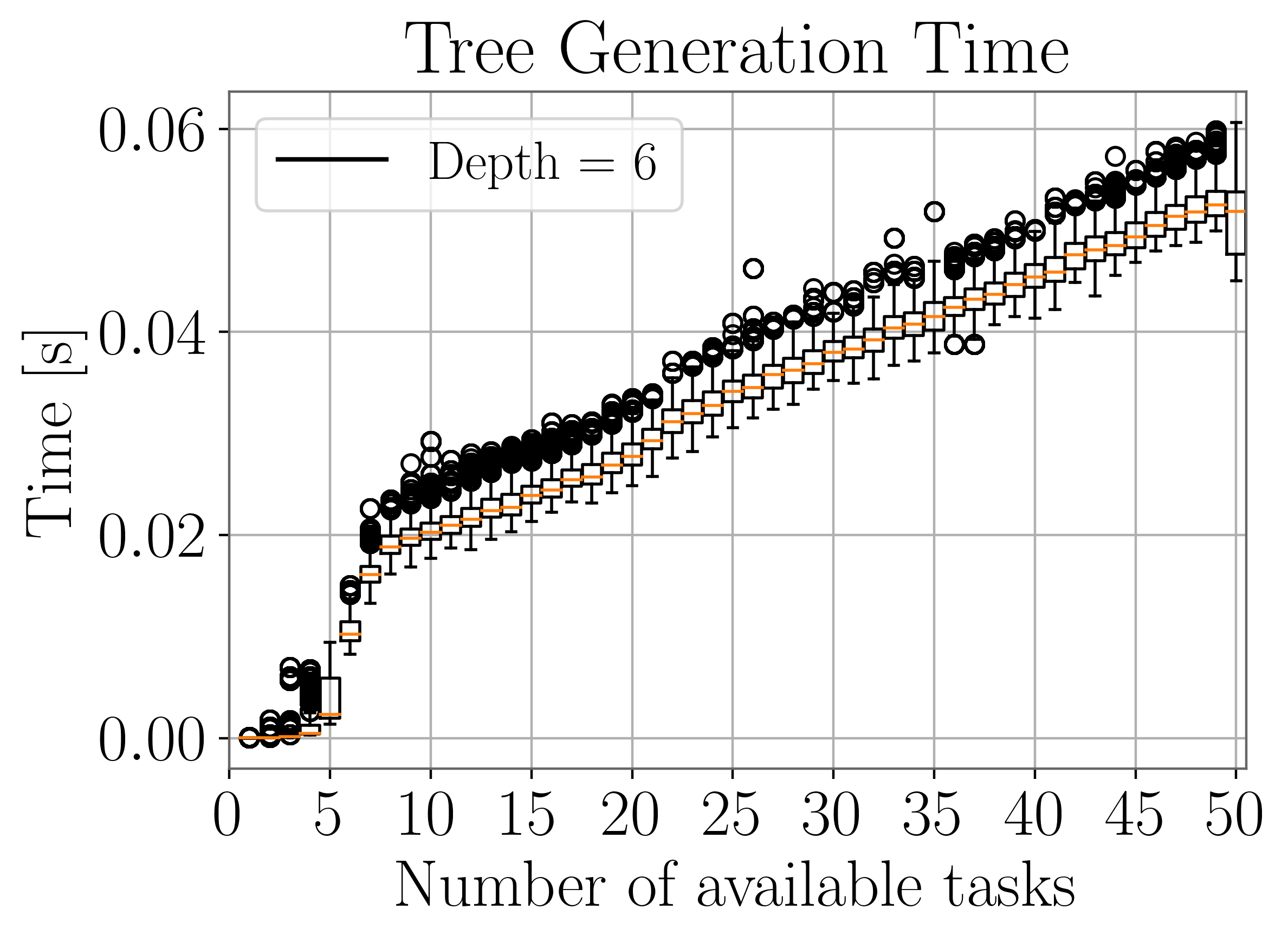}
        \caption{}
    \end{subcaptionblock}%
    \hfill
    \begin{subcaptionblock}[c]{0.25\textwidth}
        \centering
        \includegraphics[width = \textwidth]{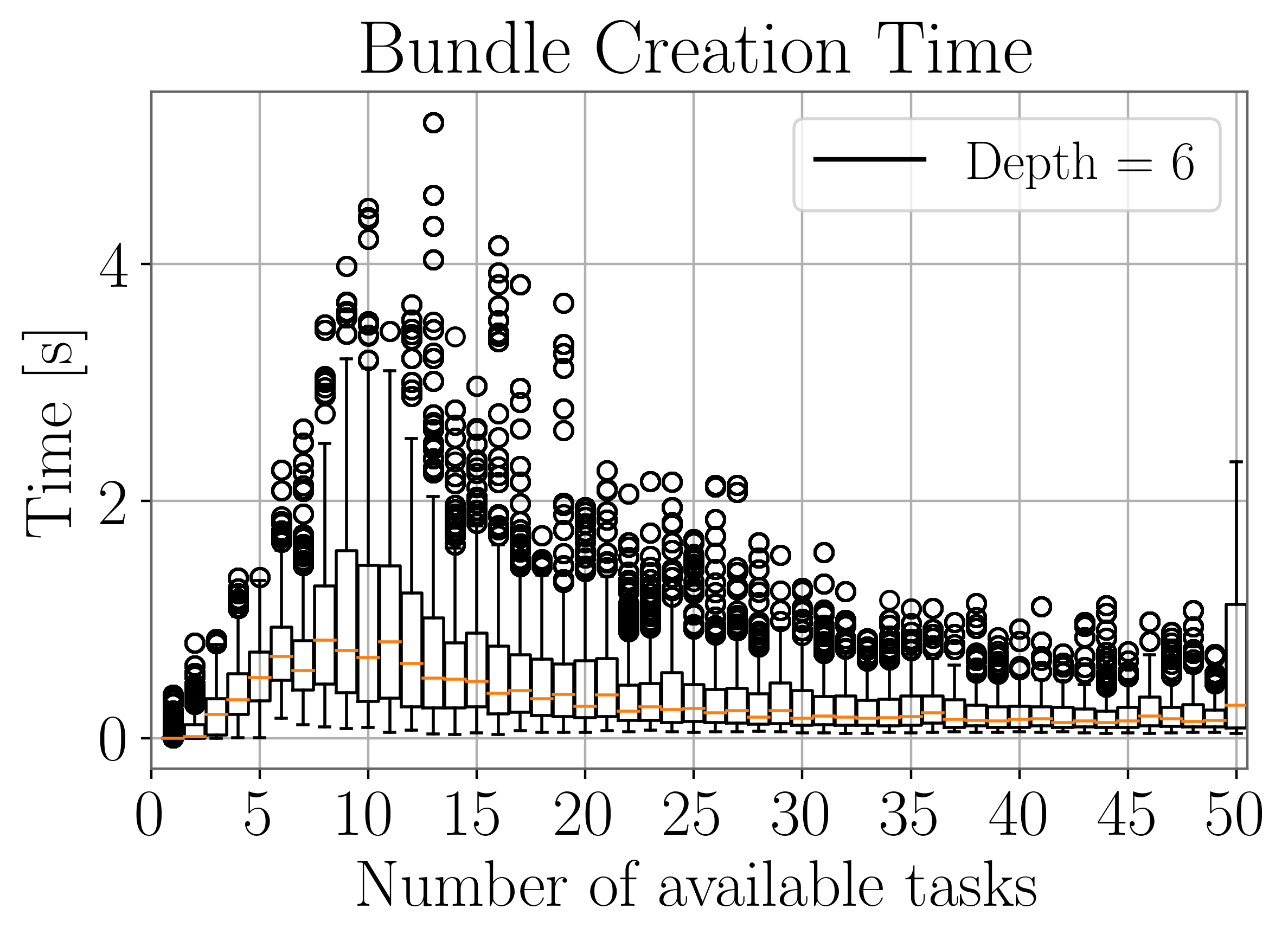}
        \caption{}
    \end{subcaptionblock}%
    \hfill
    \begin{subcaptionblock}[c]{0.25\textwidth}
        \centering
        \includegraphics[width = \textwidth]{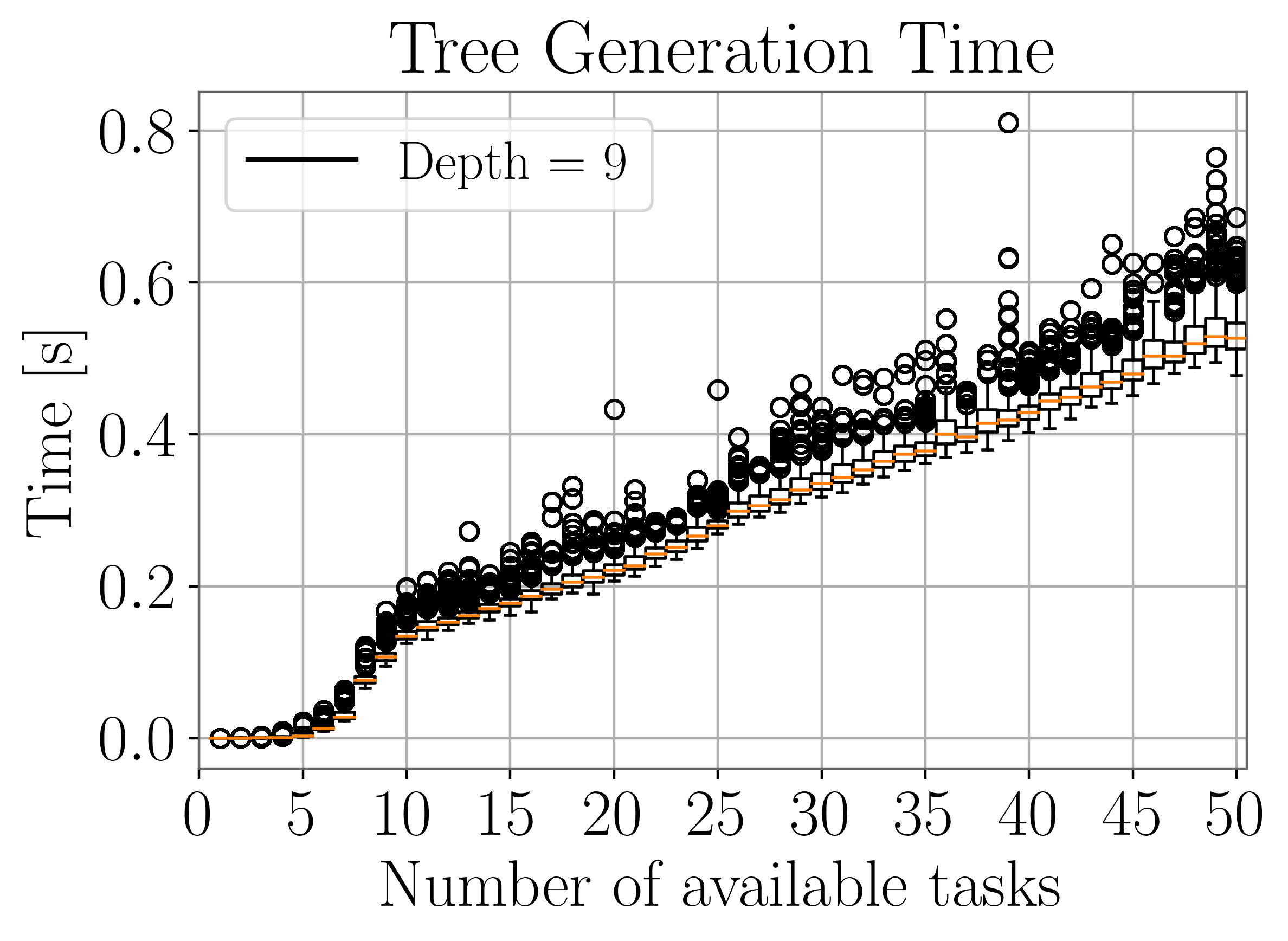}
        \caption{}
    \end{subcaptionblock}%
    \hfill
    \begin{subcaptionblock}[c]{0.25\textwidth}
        \centering
        \includegraphics[width = \textwidth]{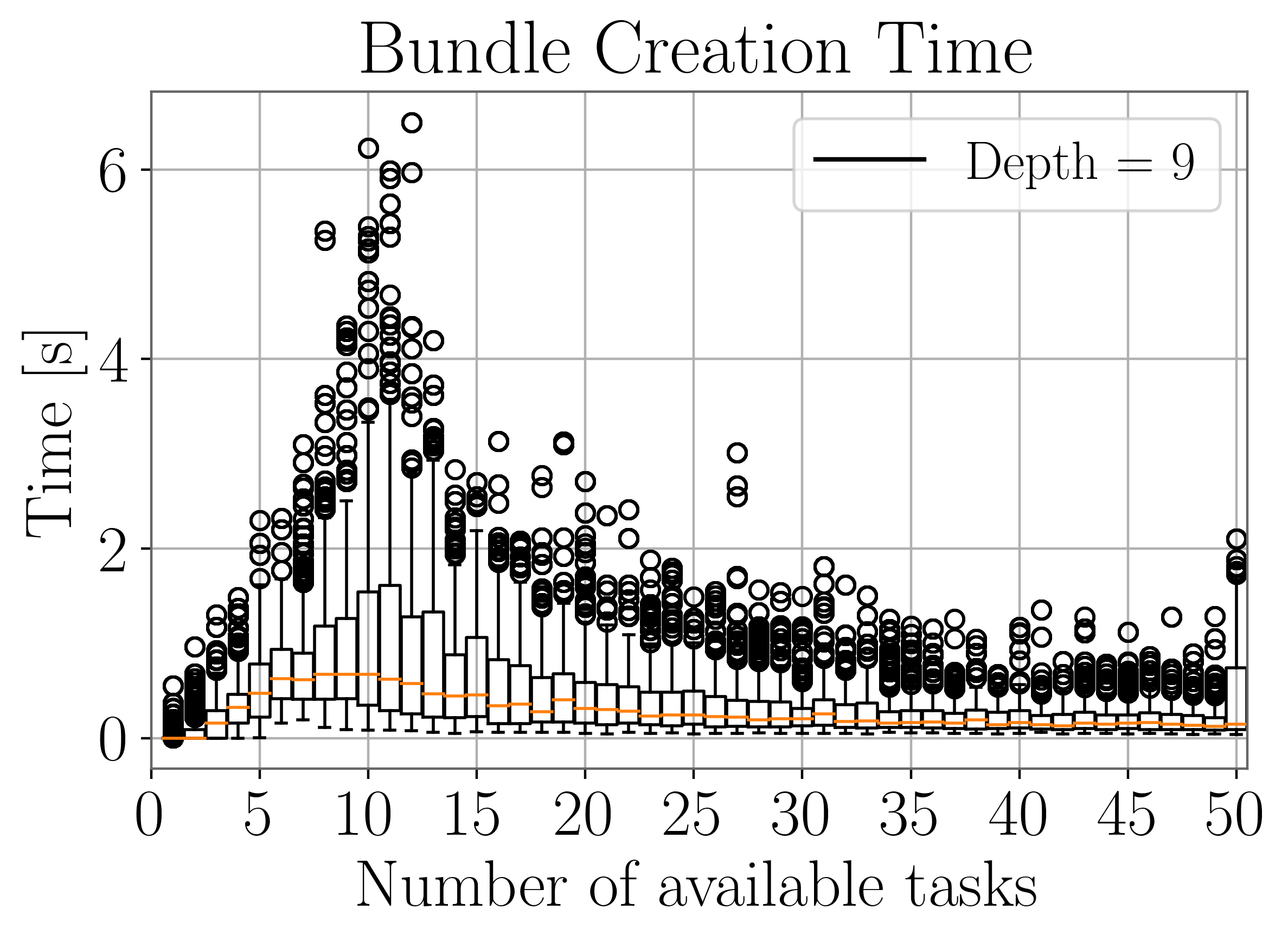}
        \caption{}
    \end{subcaptionblock}%
    
    \caption{Computation times of the proposed framework during scenario 1, map A.}
    \label{fig:computation_times_map_a}
\end{figure*}

\begin{figure*}[tb]
    \centering
    \begin{subcaptionblock}[c]{0.25\textwidth}
        \centering
        \includegraphics[width = \textwidth]{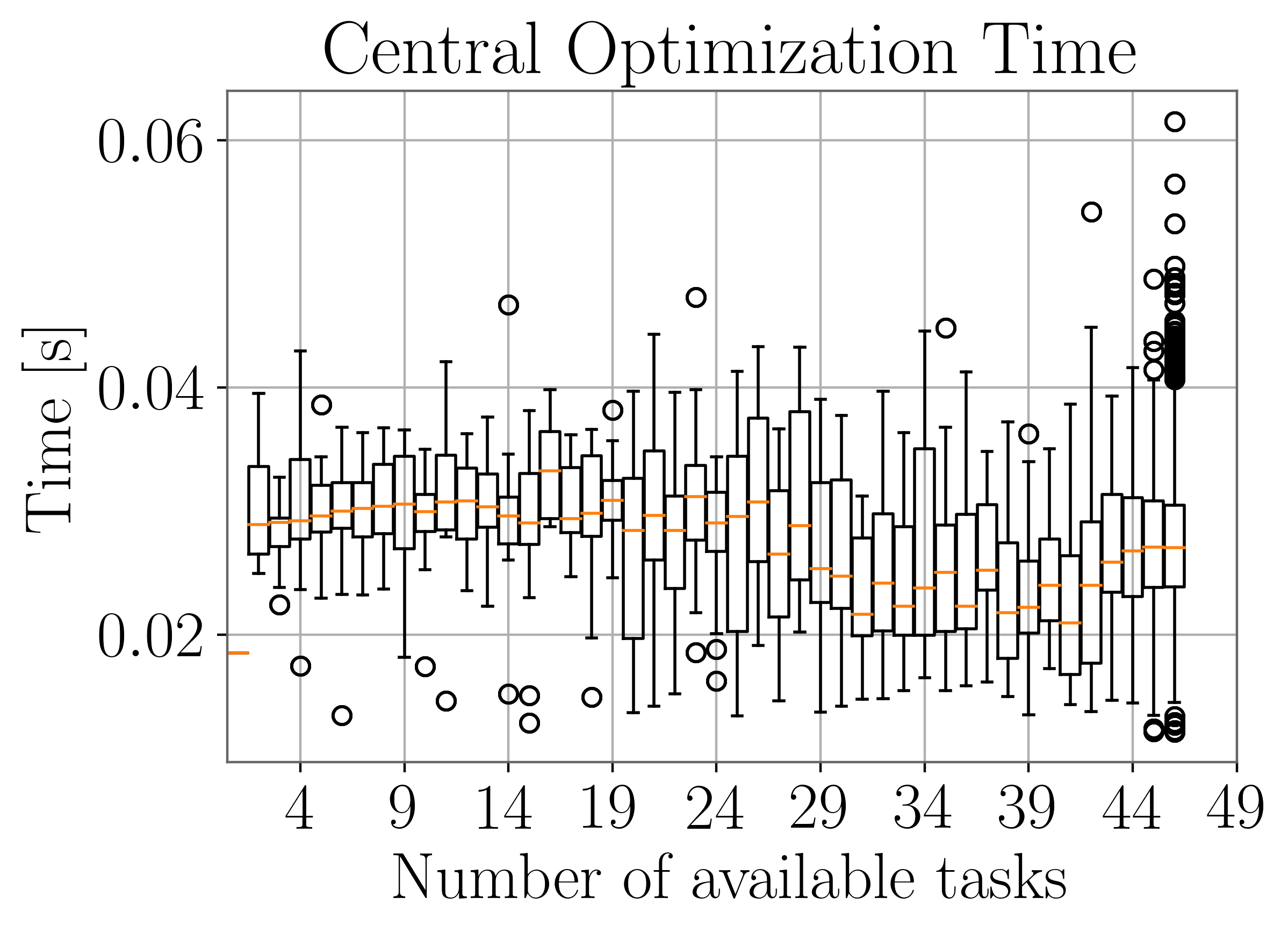}
        \caption{}
    \end{subcaptionblock}%
    \hfill
    \begin{subcaptionblock}[c]{0.25\textwidth}
        \centering
        \includegraphics[width = \textwidth]{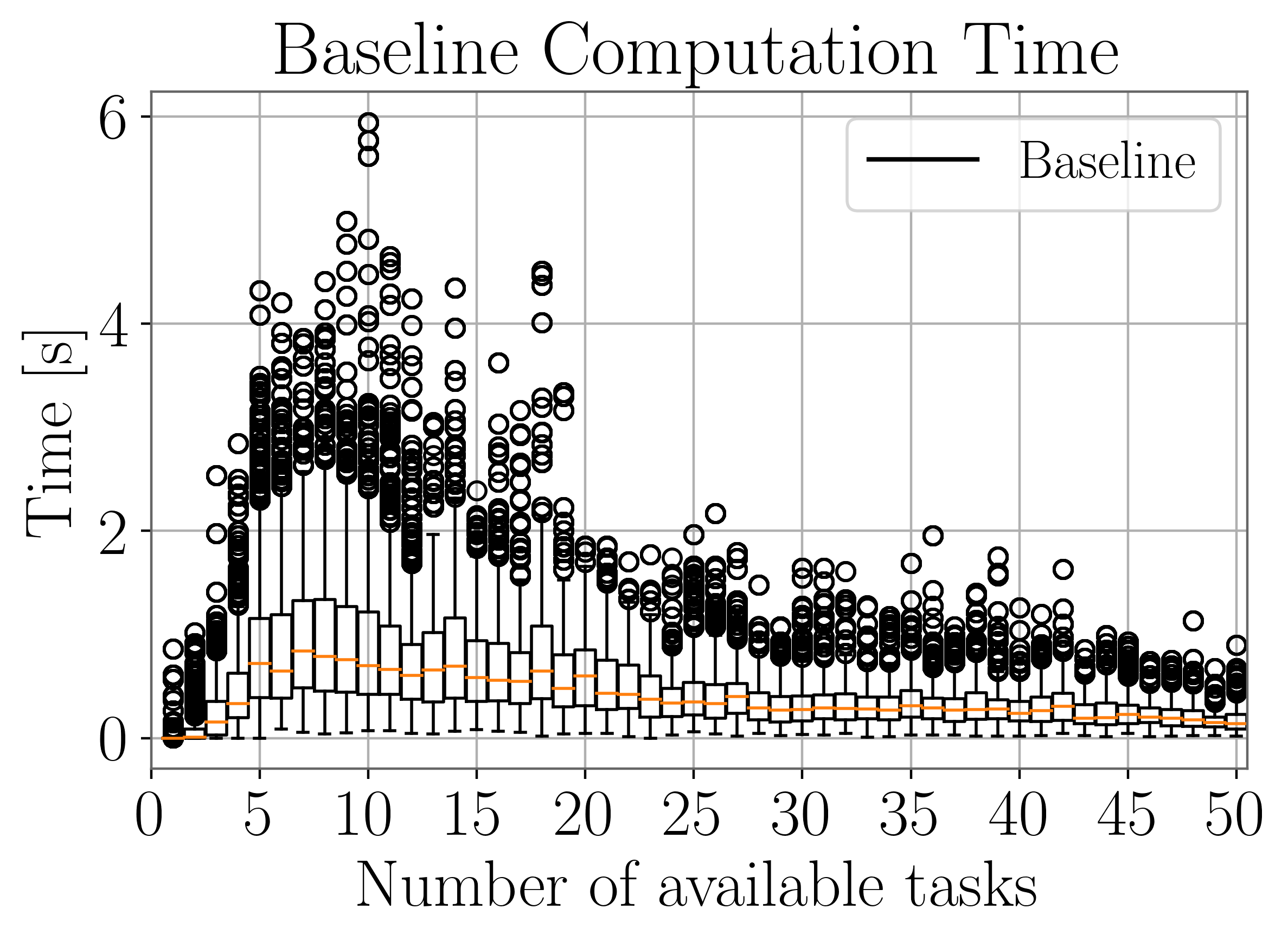}
        \caption{}
    \end{subcaptionblock}%
    \hfill
    \begin{subcaptionblock}[c]{0.25\textwidth}
        \centering
        \includegraphics[width = \textwidth]{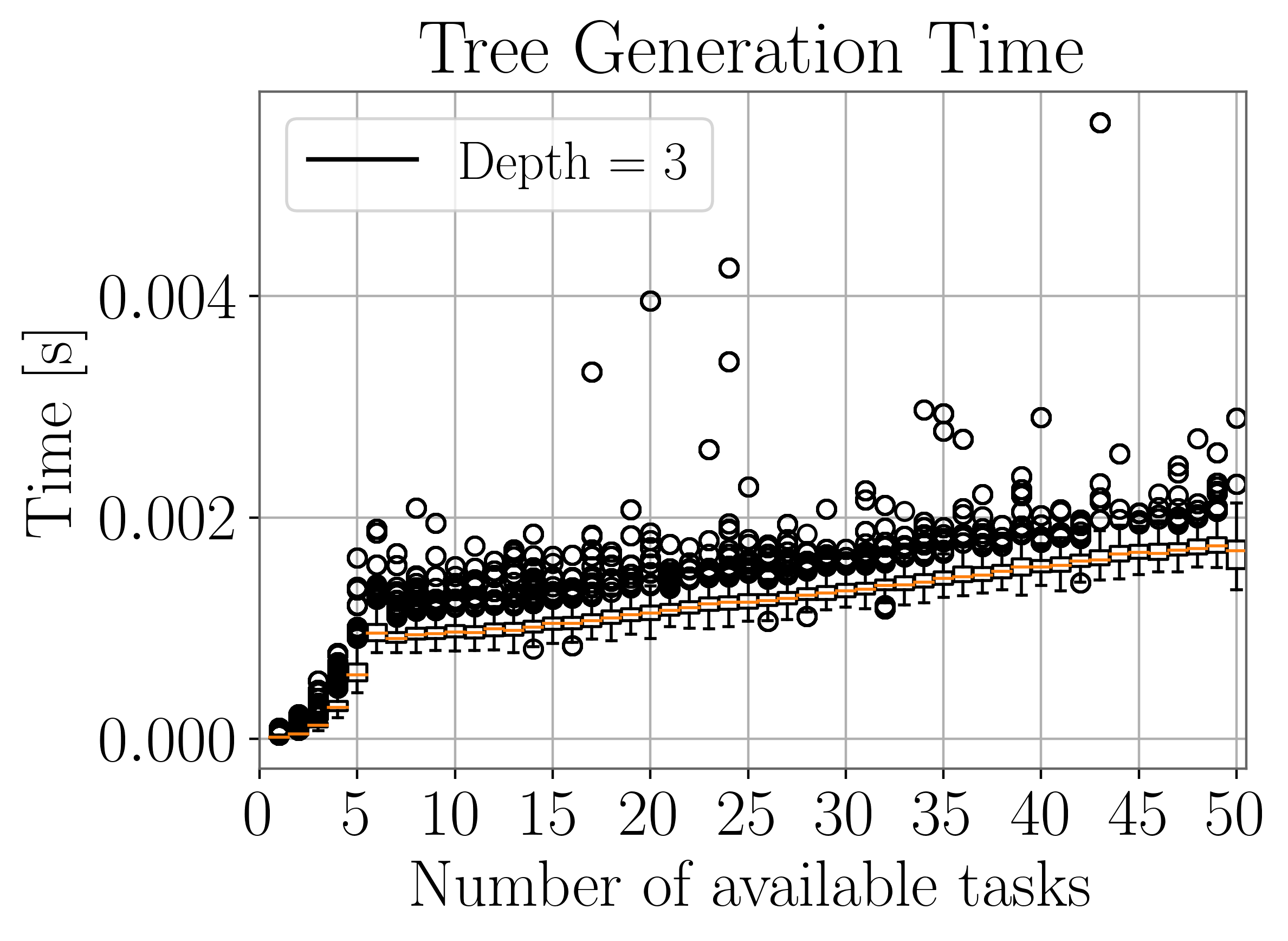}
        \caption{}
    \end{subcaptionblock}%
    \hfill
    \begin{subcaptionblock}[c]{0.25\textwidth}
        \centering
        \includegraphics[width = \textwidth]{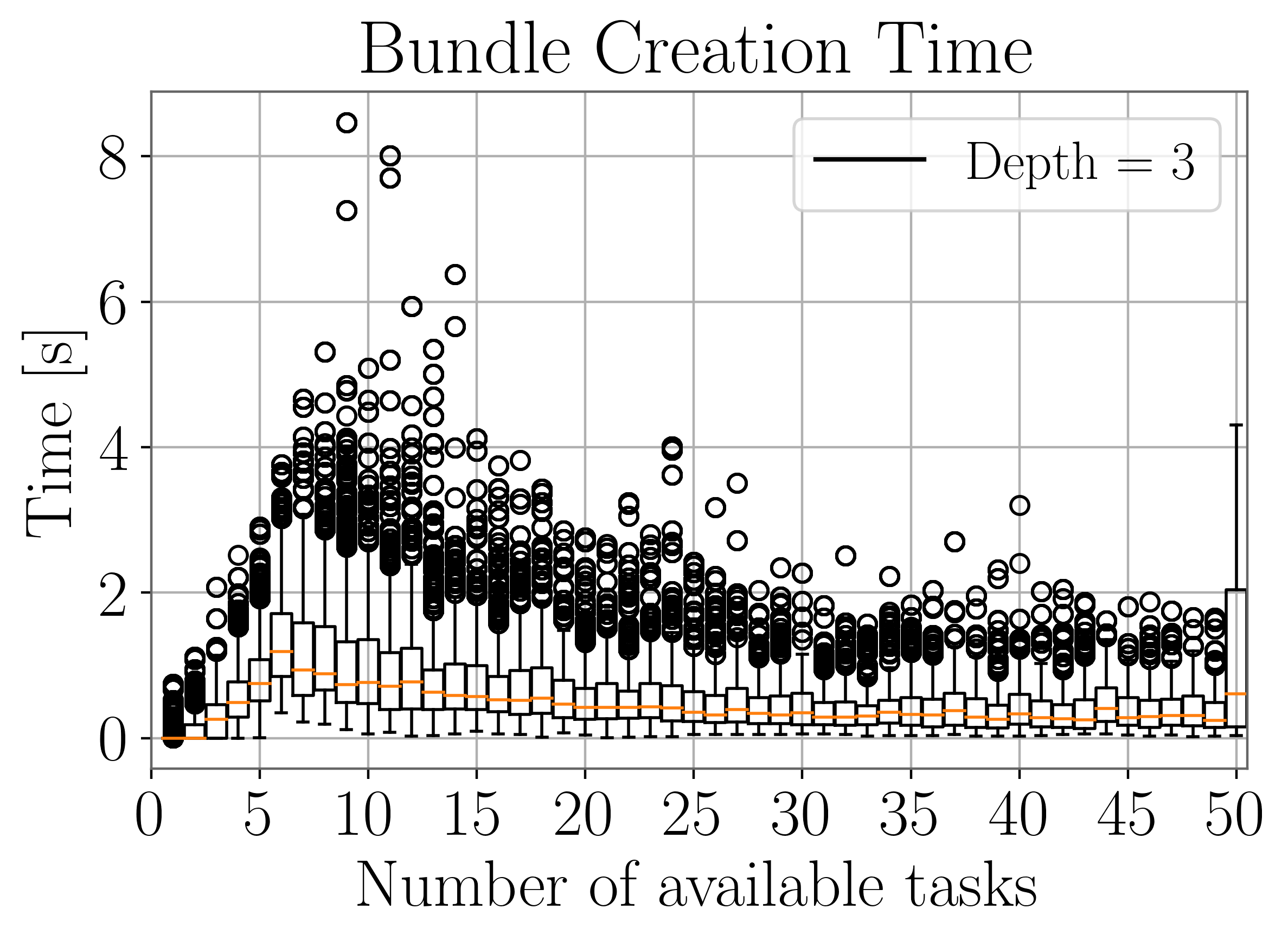}
        \caption{}
    \end{subcaptionblock}%
    \\
    \begin{subcaptionblock}[c]{0.25\textwidth}
        \centering
        \includegraphics[width = \textwidth]{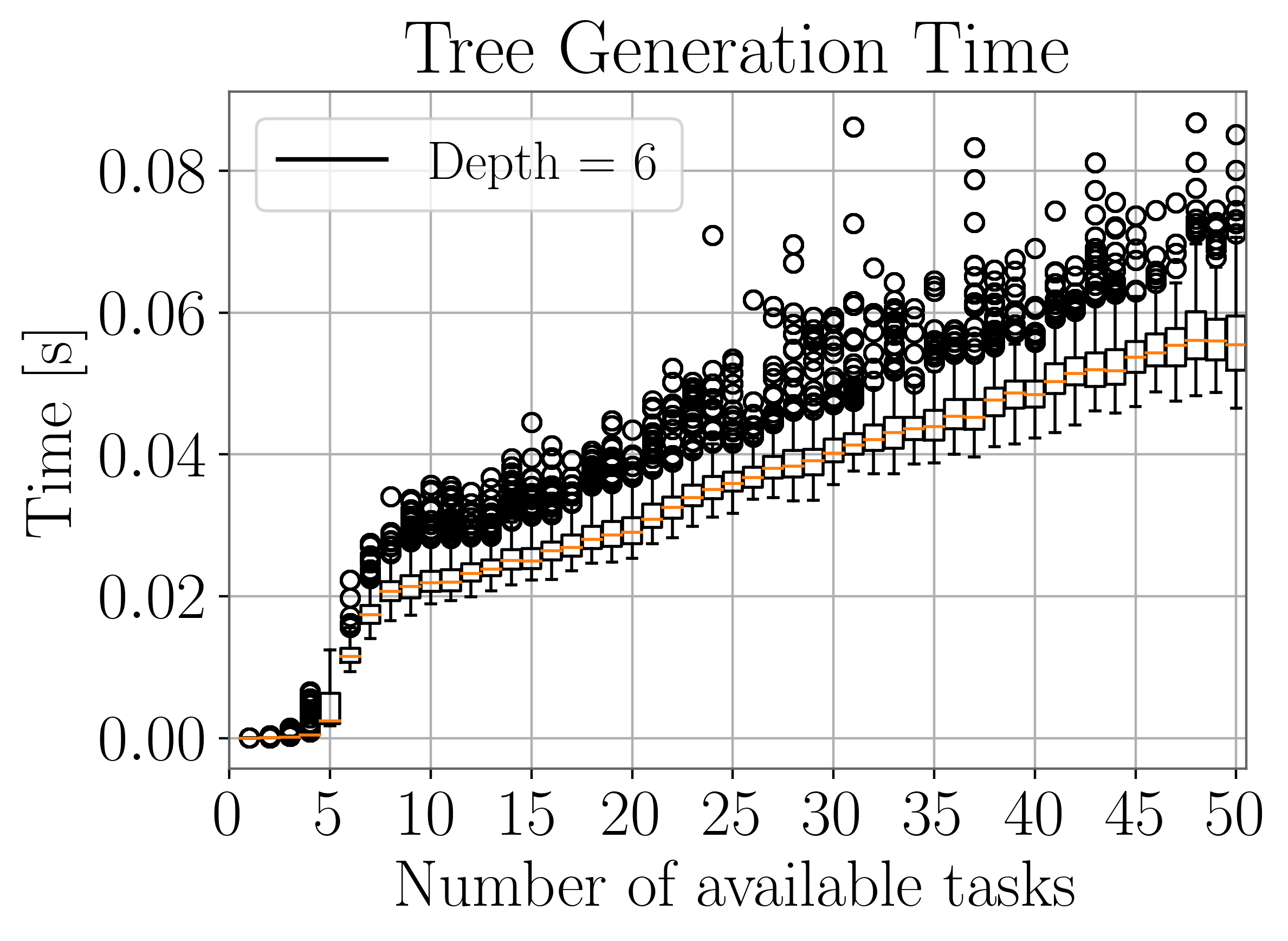}
        \caption{}
    \end{subcaptionblock}%
    \hfill
    \begin{subcaptionblock}[c]{0.25\textwidth}
        \centering
        \includegraphics[width = \textwidth]{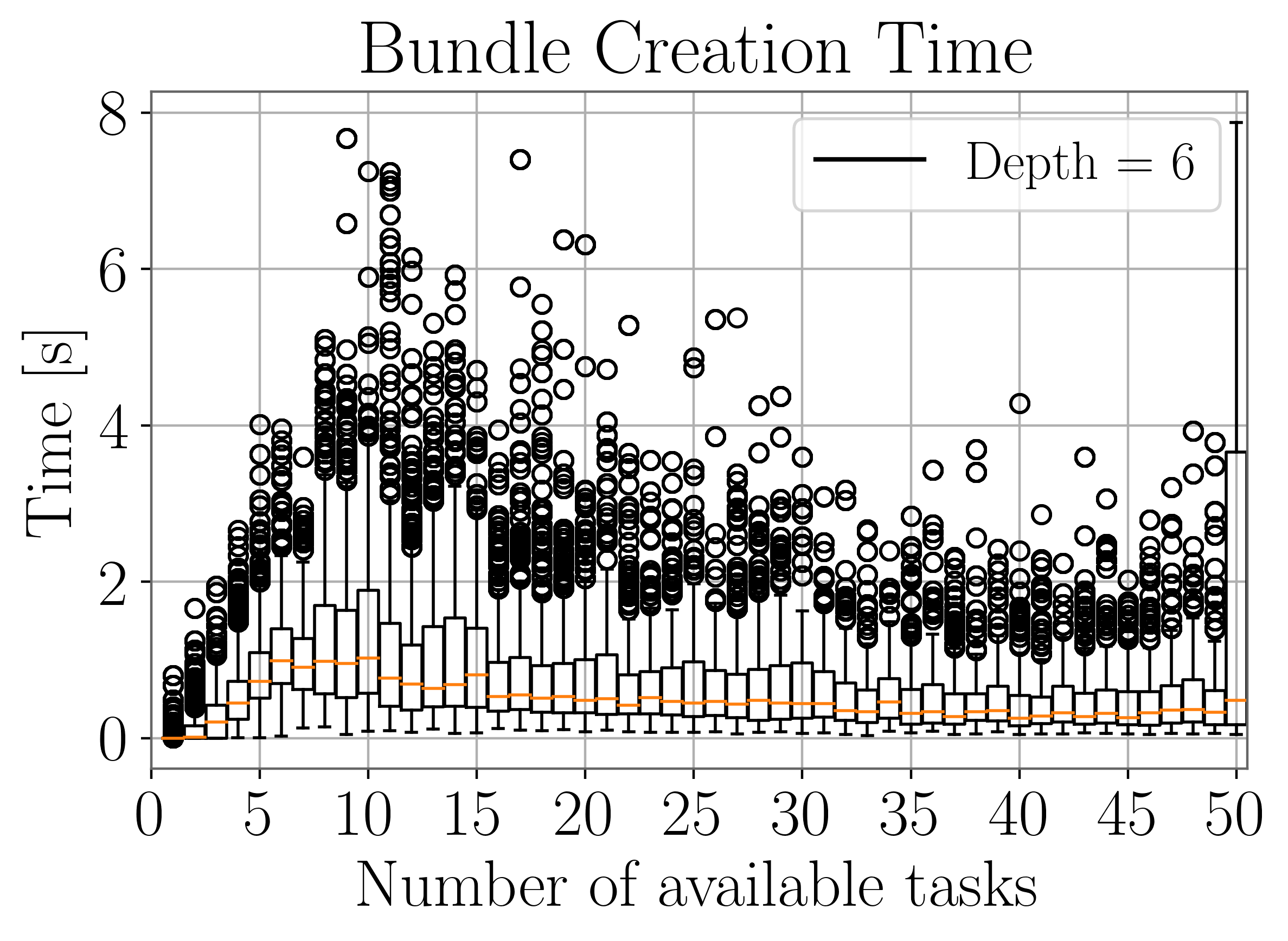}
        \caption{}
    \end{subcaptionblock}%
    \hfill
    \begin{subcaptionblock}[c]{0.25\textwidth}
        \centering
        \includegraphics[width = \textwidth]{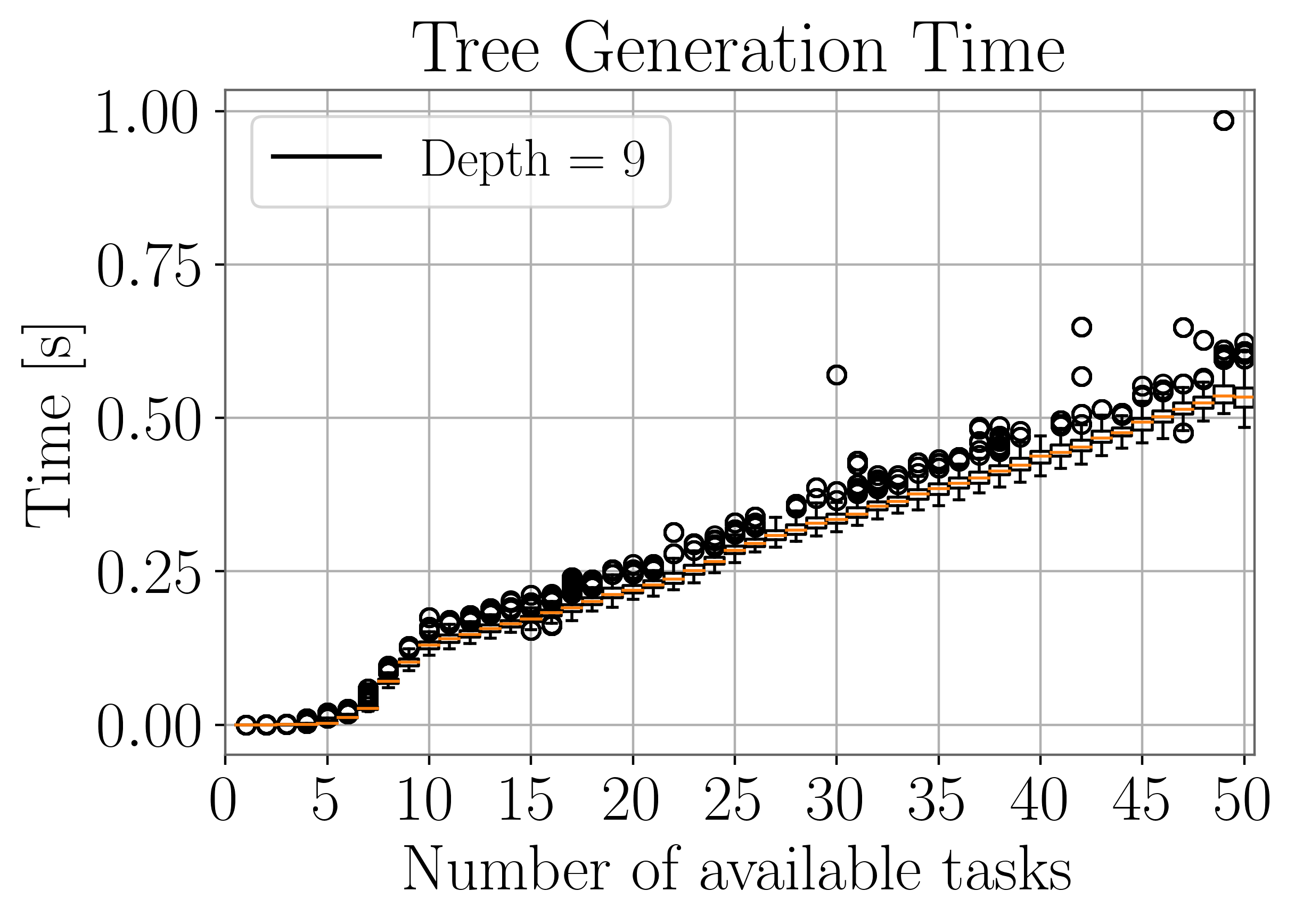}
        \caption{}
    \end{subcaptionblock}%
    \hfill
    \begin{subcaptionblock}[c]{0.25\textwidth}
        \centering
        \includegraphics[width = \textwidth]{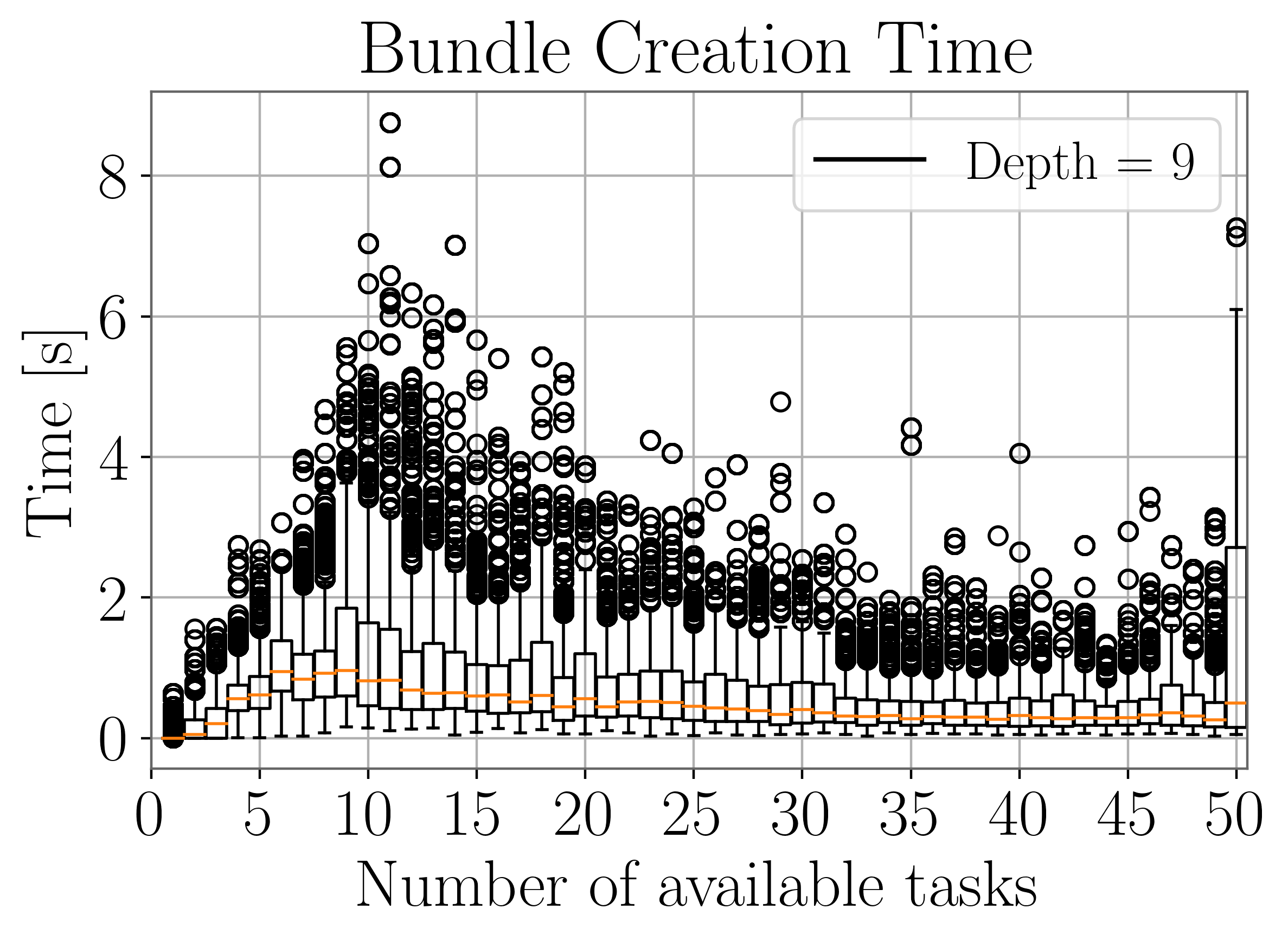}
        \caption{}
    \end{subcaptionblock}%
    
    \caption{Computation times of the proposed framework during scenario 1, map B.}
    \label{fig:computation_times_map_b}
\end{figure*}

\subsubsection{Scenario 1: Performance and Computation Efficiency:} \label{sec:scenario_1_description}
The first scenario evaluates the performance in a static setting where all tasks are introduced at initialization. A fixed set of \(50\) tasks are uniformly randomly distributed across map A, B and C. A fleet of \(4\) agents, with static starting locations, are deployed to complete all tasks. This scenario isolates the computational efficiency of the proposed two-stage bid generation process. The time required for (1) the candidate tree generation and (2) the high-fidelity cost estimation is presented in Fig. \ref{fig:computation_times_map_a} and \ref{fig:computation_times_map_b}.
The overall execution performance, the total time to complete all available tasks as shown in Fig. \ref{fig:total_steps_static}, is compared to a fully reactive baseline where agents utilizes a greedy approach where only the \(6\) nearest tasks are considered as bids (as described in \cite{10566142}).

\subsubsection{Scenario 2: Dynamic Task Insertion:}
The second scenario introduces temporal uncertainty to evaluate the proposed methods responsiveness and the performance of the task execution in a scenario where the mission is abruptly canceled. The fleet size, and agent starting locations, remains the same as in scenario 1 but the tasks are now introduced randomly during the mission execution (with a \(10 \%\) probability of adding a task at every time step, with a maximum number of \(50\) available tasks) and the mission is abruptly aborted after \(600\) time steps. The primary metric evaluated here is the rate of task completion, illustrated in Fig. \ref{fig:total_steps_dynamic} as the number of completed tasks until the mission is aborted, compared to the same fully reactive approach explained in section \ref{sec:scenario_1_description}. The key idea is that the proposed method should try to seek out 'hot spots` of the available tasks and therefore complete more tasks within the allowed execution time.

\begin{figure*}[h!]
    \centering

    \begin{subcaptionblock}[c]{0.247\textwidth}
        \centering
        \includegraphics[width = \textwidth]{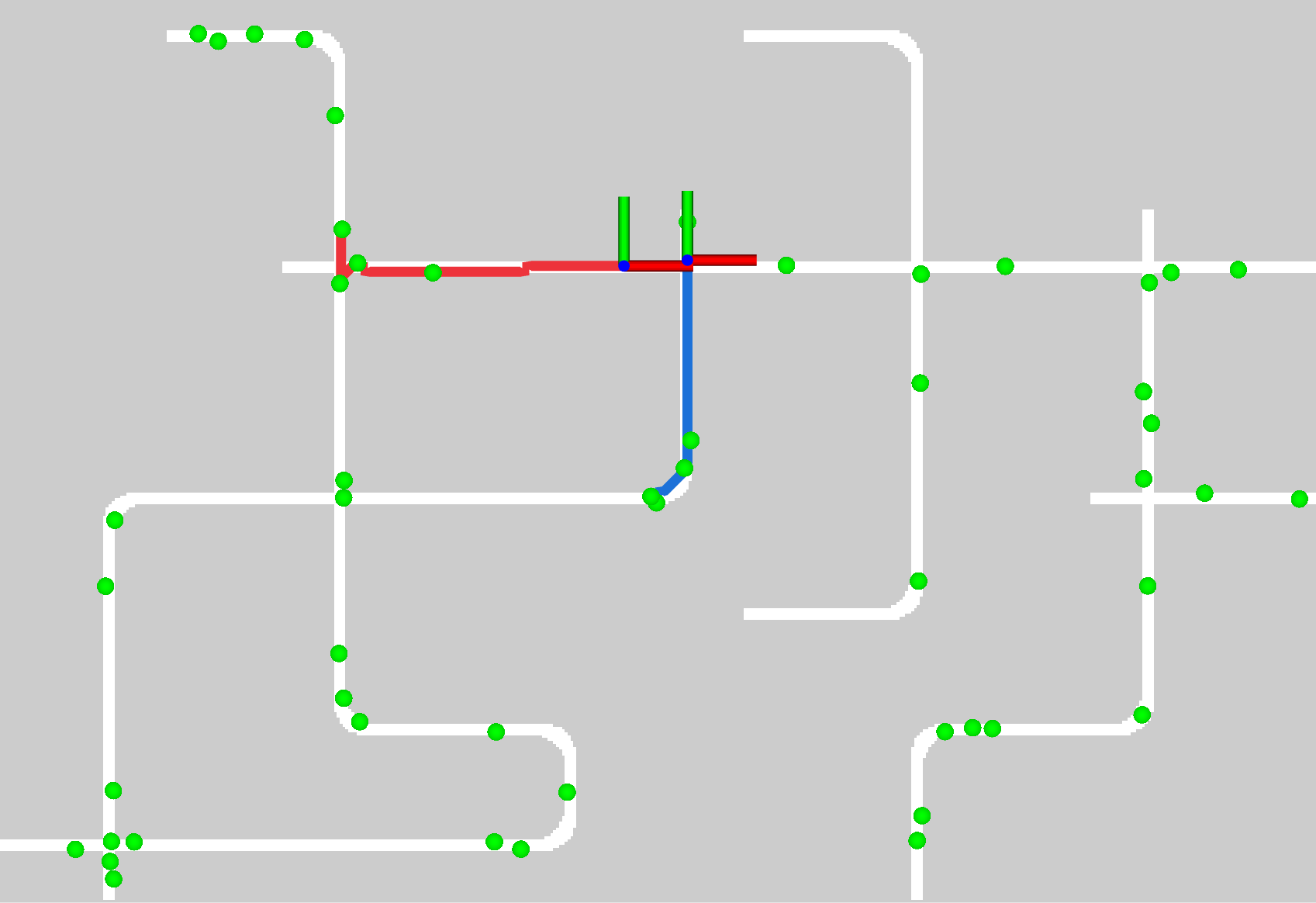}
        \vspace{0.5pt}
        \caption{t = 0}
    \end{subcaptionblock}%
    \hfill
    \begin{subcaptionblock}[c]{0.247\textwidth}
        \centering
        \includegraphics[width = \textwidth]{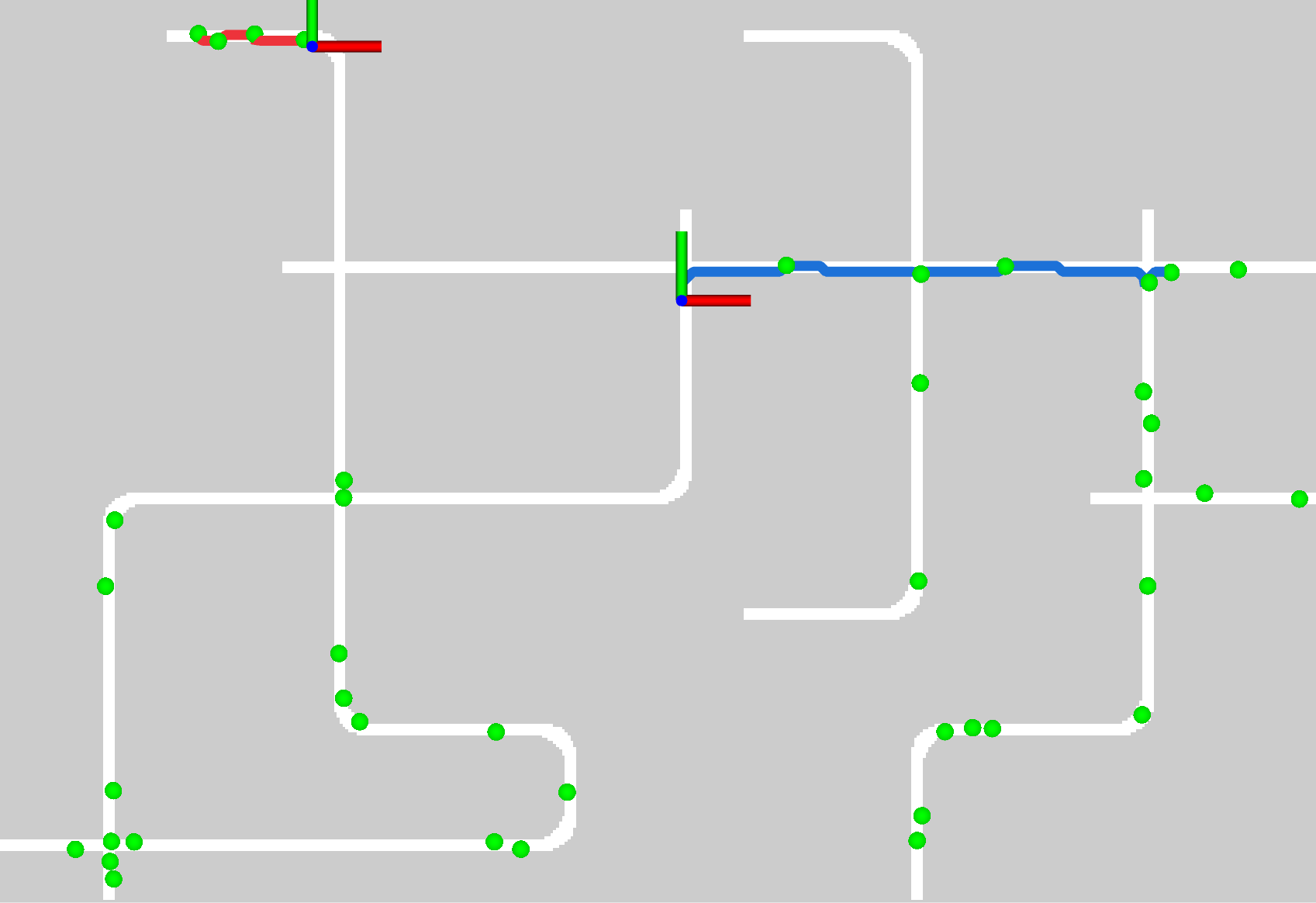}
        \vspace{0.5pt}
        \caption{t = 100}
    \end{subcaptionblock}%
    \hfill
    \begin{subcaptionblock}[c]{0.247\textwidth}
        \centering
        \includegraphics[width = \textwidth]{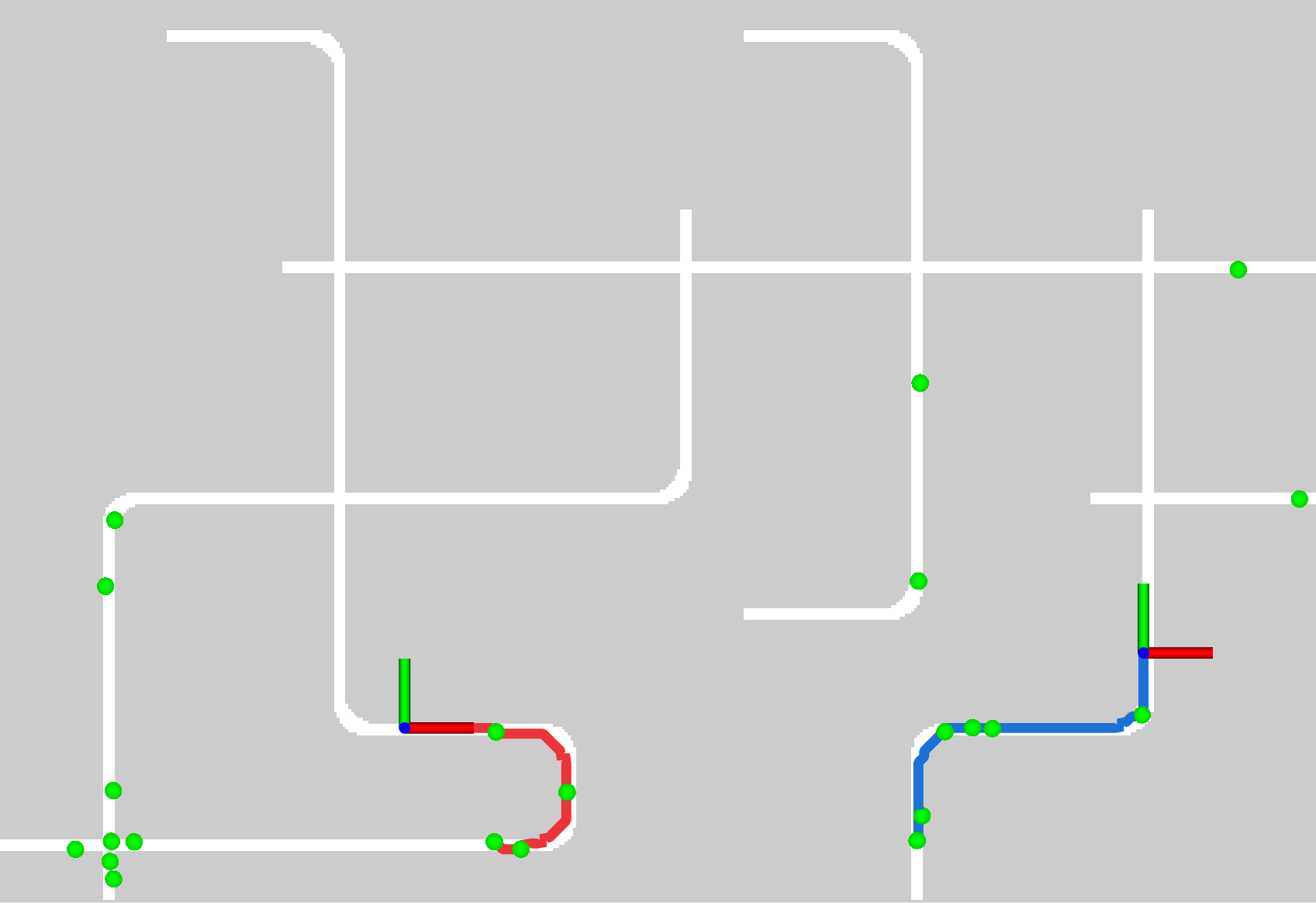}
        \vspace{0.5pt}
        \caption{t = 275}
    \end{subcaptionblock}%
    \hfill
    \begin{subcaptionblock}[c]{0.247\textwidth}
        \centering
        \includegraphics[width = \textwidth]{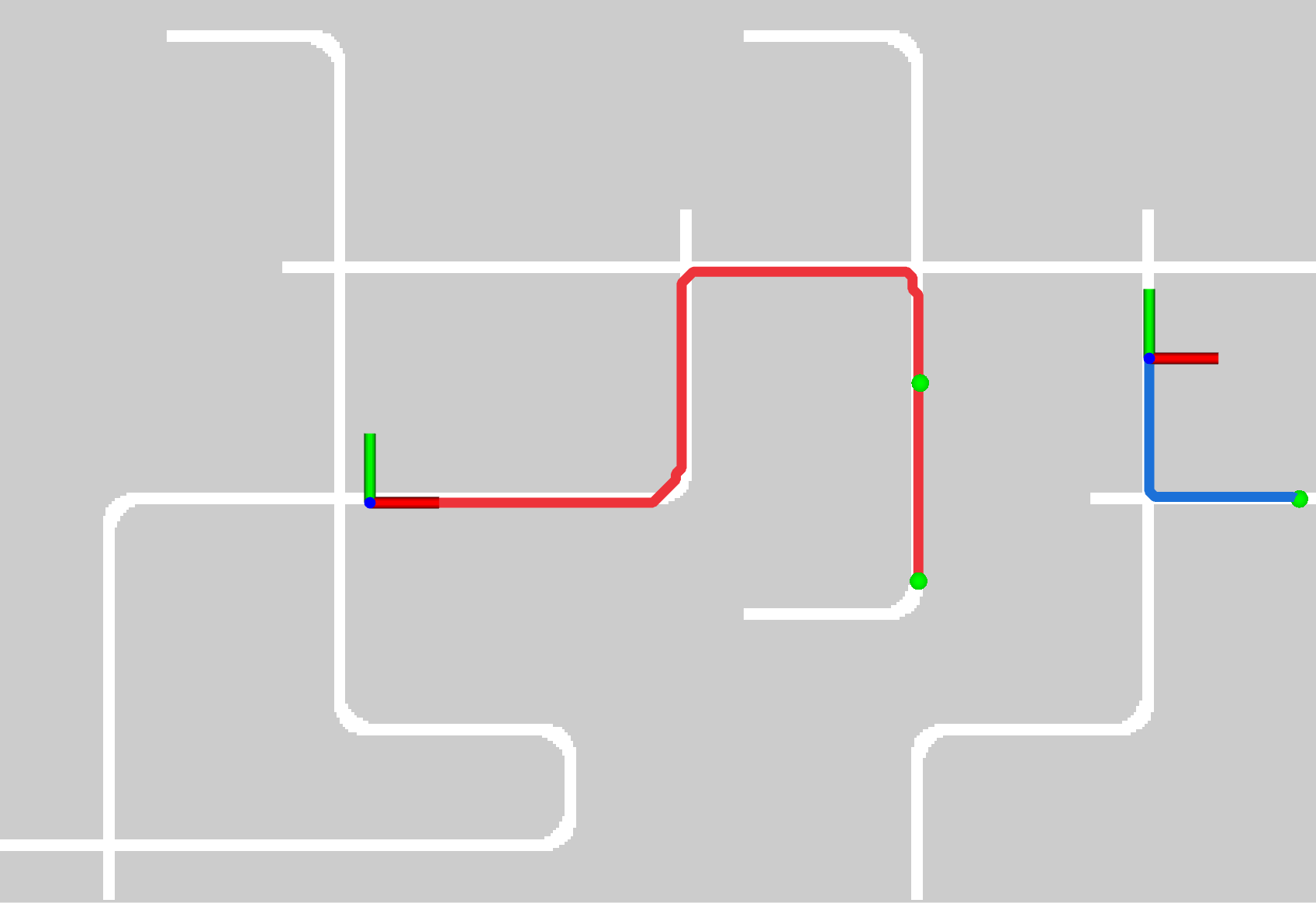}
        \vspace{0.5pt}
        \caption{t = 520}
    \end{subcaptionblock}%
    \caption{Snapshots from scenario 3. The available tasks are represented as green dots and the current path for the two agents are shown as a red and blue line respectively. 
    }
    \label{fig:illustrative_scenario_snapshots}
\end{figure*}

\subsubsection{Scenario 3: Qualitative Tunnel Inspection:}

The final scenario replicates a representative real-world use case where a collaborative inspection mission is carried out in a tunnel environment. A long tunnel map is used during this scenario where euclidean distance is a poor heuristic for actual travel cost. In this scenario, \(50\) tasks, representing inspection points along the tunnel walls, are introduced and a fleet of two agents are utilized to provide a qualitative assessment of the task distribution and bundle formation in a complex environment. A series of snapshots showing the allocation state at various time steps during the mission execution is shown in Fig. \ref{fig:illustrative_scenario_snapshots}. These snapshots illustrate how the agents partition the available tasks and how the candidate tree successfully navigates the topology and provides good task bundles for the agents.

\section{Discussion} \label{sec:discussion}

The results from Scenario 1 (static allocation) empirically validates the increased performance of the proposed method. That a two-stage, multi-fidelity bundle generation process can bridge the gap between a fast but inaccurate heuristic and a slow but precise planner for accurate cost estimation. Despite the aggressive pruning of the available bundles, the proposed method manages to consistently perform \(\sim 14-18\%\) better on average in completing all the available tasks compared to the reactive baseline even though the computation time remains similar to the baseline. This gain is mainly attributed to the frameworks ability to capture task synergies and recognizing that a cluster of tasks should be executed by one agent rather than split among all available agents.

A critical design choice in the proposed framework is the reliance on a euclidean distance heuristic during the candidate tree generation. We acknowledge that in non-convex environments, such as the cave- and tunnel inspired maps, euclidean proximity often diverges from the true travel cost. We deliberately trade initial heuristic accuracy for computation speed. The risk of pruning valid candidates is minimized by retaining multiple diverse branches at each depth to preserves a sufficiently broad solution space.

Scenario 2 also shows an improvement in performance. Here, considering dynamic tasks without a priori information, the proposed method outperforms the reactive baseline by \(\sim 10-17\%\) on average. The introduction of an abrupt mission termination after a fixed time shows that the proposed method consistently seeks out `hot spots' of tasks to increase the overall system performance.

Scenario 3 (tunnel inspection use-case) provides a qualitative verification of the frameworks applicability to a motivating use cases. It is shown that the proposed method is able to handle non-convex environments where the euclidean distance heuristic might give cost estimates very far from the true costs. The visualization video reveals an emergent spatial partitioning of the environment. Where the two agents initially focuses on the most prominent clusters of tasks and spread out to different parts of the environment. A video showing the execution of scenario 3 is provided at \url{https://youtu.be/j5uhG0DAJR0} .

\section{Conclusions and Future Work} \label{sec:conclusions}

This paper presented a scalable reactive framework for multi-robot task allocation designed for dynamic environments where estimating execution costs is computationally expensive. By decomposing the problem into a distributed, two-stage bundle generation process and a centralized auction-based set-packing optimization, we effectively bridge the gap between combinatorial market-based approaches and robotic path planning. The introduction of a multi-fidelity candidate tree allows agents to explore a large space of potential tasks using a cheap heuristics, applying expensive path planning only to the most promising candidates through a best-first search. Furthermore, simulation results confirm that this approach delivers higher quality solution compared to fully reactive auction based task allocation framework.

Future work will focus on further validating the proposed framework by coupling the candidate tree generation together with the best-first cost estimation for a more efficient framework and showing how the approach could be extended to handle situations where budget constraints and time-constraints for tasks are considered.



\bibliography{bib}             
                                                   







\end{document}